\documentclass{article}
\usepackage{iclr2026_conference,times}

\usepackage{amsmath,amsfonts,bm}

\def\eqref#1{equation~\ref{#1}}

\def\1{\bm{1}}

\DeclareMathAlphabet{\mathsfit}{\encodingdefault}{\sfdefault}{m}{sl}
\SetMathAlphabet{\mathsfit}{bold}{\encodingdefault}{\sfdefault}{bx}{n}

\usepackage{hyperref}
\usepackage{url}
\usepackage{booktabs}
\usepackage{graphicx}
\usepackage{amsmath}
\usepackage{multirow}
\usepackage{xcolor}
\usepackage{subcaption}
\usepackage{float}
\usepackage{algorithm}
\usepackage{algpseudocode}
\usepackage{wrapfig}
\usepackage{colortbl}
\usepackage{rotating}
\usepackage{array}
\definecolor{cBest}{HTML}{9CC3E4}
\definecolor{cSecond}{HTML}{DCEAF6}
\definecolor{cPos}{HTML}{63B36B}
\definecolor{cNeg}{HTML}{D98A8A}
\definecolor{cStd}{HTML}{8A8F98}
\newcommand{\std}[1]{\textcolor{cStd}{\scriptsize$_{\pm#1}$}}
\renewcommand{\arraystretch}{1.15}

\title{One Rewrite to Fix Them All?\\Type-Aware Repair Allocation for Text-to-Image Prompt Optimization}

\author{
\textbf{Haoyue Liu}\textsuperscript{1}
\quad
\textbf{Xiaoyu Ma}\textsuperscript{1}
\quad
\textbf{Ye Chen}\textsuperscript{2}
\quad
\textbf{Shuguang Cui}\textsuperscript{1,3}
\quad
\textbf{Xiaoying Tang}\textsuperscript{1,3,\ensuremath{\dagger}}
\\[0.6em]
\textsuperscript{1}
School of Science and Engineering,
The Chinese University of Hong Kong, Shenzhen 518172, China
\\
\textsuperscript{2}
XJTU-POLIMI Joint School, Xi'an Jiaotong University, Xi'an 710049, China
\\
\textsuperscript{3}
Shenzhen Future Network of Intelligence Institute (FNii-Shenzhen)
}

\iclrfinalcopy

\begin{document}
\maketitle

\begingroup
\renewcommand{\thefootnote}{\fnsymbol{footnote}}
\footnotetext[2]{Corresponding author.}
\endgroup

\fancyhead{}

\begin{abstract}
Text-to-image (T2I) generators often fail to follow their prompts faithfully, producing wrong counts, swapped attributes, ambiguous relations, and illegible text. Prompt optimization repairs such failures by rewriting the user prompt, requiring no generator retraining, and has yielded promising results. However, existing optimizers absorb heterogeneous failures into one uniform prompt expansion, even though each calls for different repair language. We formulate semantic prompt optimization as \emph{atomic repair allocation}: each failed proposition is routed to a type-conditioned repair operator before the resulting local constraints are compiled into one executable prompt. We instantiate this formulation in the training-free Type-Aware Repair Allocation (TARA) framework, which separates diagnosis, allocation, compilation, and a semantic repair gate---an accept-or-revert controller over exactly one prescribed repair that prevents semantic regressions. Extensive experiments on DSG and TIFA across four frozen generators demonstrate that TARA achieves the best semantic accuracy in all eight benchmark--generator cells, improving over VisualPrompter by $+5.6$/$+2.6$ points on DSG/TIFA, while maintaining image quality and running fastest in our matched local setting ($16.0$s versus $20.0$s per prompt).
\end{abstract}

\section{Introduction}

Text-to-image (T2I) generation has become a default way to turn language into pictures~\citep{ho2020denoising,rombach2022high,blackforestlabs2024flux,chen2025januspro}, yet making an image actually satisfy its prompt is still fragile: modern generators routinely drop a requested object, miscount, swap an attribute, scramble a relation, or render text no one can read. Prompt optimization has therefore emerged as a practical remedy that requires no generator retraining. Keyword-driven rewriters such as Promptist~\citep{hao2023optimizing}, BeautifulPrompt~\citep{cao2023beautifulprompt}, NeuroPrompts~\citep{rosenman2024neuroprompts}, and TIPO~\citep{yeh2024tipo} increase visual appeal, and recent visual-feedback methods such as VisualPrompter~\citep{wu2026visualprompter} decompose the prompt into atomic propositions and detect missing concepts with a VLM, targeting semantic faithfulness directly.

\begin{figure}[h!]
\centering
\includegraphics[width=0.78\textwidth]{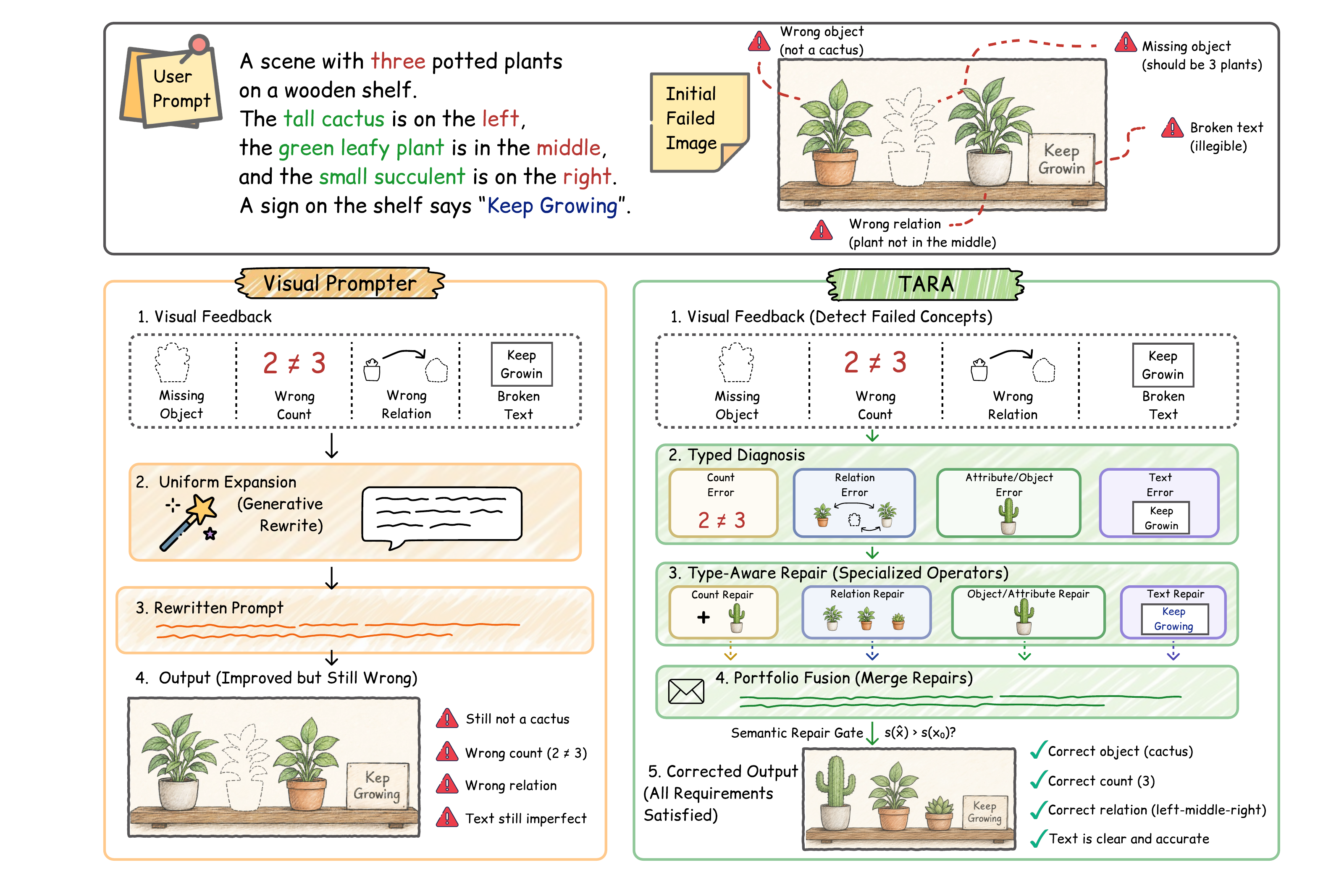}
\caption{Type-agnostic expansion (VisualPrompter, left) versus atomic repair allocation (TARA, right). One uniform expansion leaves several fine-grained failures unresolved, whereas TARA routes each failure to a type-conditioned repair within the same single re-generation budget.}
\label{fig:motivation}
\end{figure}

Despite this progress, existing optimizers share an unavoidable drawback: \emph{whatever the failure, the prompt is repaired by a single uniform expansion}. A missing object, a wrong count, a broken relation, and illegible text are handled by the same enrichment rule, so the repair seldom matches the failure; VisualPrompter, for instance, uses atomic feedback to decide \emph{what} is missing, but still applies a largely type-agnostic strategy when deciding \emph{how} to rewrite it. As illustrated in Figure~\ref{fig:motivation}, on a prompt that jointly demands a count, an attribute, a layout, and a legible sign, one such expansion leaves several errors standing. Notably, \emph{different failures call for different repair language}: a missing object must be made salient, a wrong count needs explicit cardinality and separation, a spatial relation needs an unambiguous layout, and rendered text needs the exact string with legible typography. Under our formulation, uniform expansion is a degenerate allocation policy with one type-invariant shared operator across failure types. Rather than using atomic categories only to describe what failed, TARA operationalizes them as routing variables that determine how each failed proposition is repaired.

To tackle the above problem, we rethink how visual feedback should drive prompt rewriting and address the following core question:
\begin{center}
\emph{Can atomic visual feedback be used not merely to trigger a rewrite, but to route heterogeneous failures to specialized repairs and compile them into one reliable prompt under a single-regeneration budget?}
\end{center}

Hence, we formulate semantic prompt optimization as \emph{atomic repair allocation} and instantiate it with the \emph{Type-Aware Repair Allocation} (TARA) method. TARA follows four explicit stages: \emph{Diagnose} which atomic propositions fail; \emph{Allocate} each failure to a type-conditioned repair operator; \emph{Compile} the resulting local constraints into one concise executable prompt through a text-only portfolio-and-fusion step; and \emph{Adopt} the single repaired image only when a semantic repair gate deems it a reliable replacement. This design separates decisions that existing prompt optimizers conflate or leave uncontrolled: what went wrong, how each failure should be repaired, how heterogeneous local repairs should coexist in one prompt, and whether the result should replace the original output. The repair budget is a single additional image generation, matched to prior visual-feedback optimizers; TARA needs no task-specific training labels, generator finetuning, or white-box access, and can be applied out of the box across diverse frozen T2I generators. We summarize the contributions of this paper as follows:
\begin{itemize}
    \item We formulate visual-feedback prompt optimization as \emph{atomic repair allocation}. Rather than using atomic feedback only to trigger one global expansion, TARA treats each failed proposition as an intervention unit and assigns it a type-conditioned local repair; uniform expansion is the single-operator special case.
    \item We instantiate this formulation with TARA, a training-free framework that compiles heterogeneous local repairs into one executable prompt under a single-regeneration budget and uses a semantic repair gate to prevent the prescribed repair from introducing regressions.
    \item Across DSG and TIFA, four generators, and three seeds, TARA leads all eight benchmark--generator cells. Its gains persist across semantic types and independent evaluators, while controlled ablations isolate the explicit typed-expansion branch, repair compilation, recovery, and adoption under a matched image budget.
\end{itemize}

\section{TARA}

The discussion above motivates a four-stage view of prompt optimization: \emph{Diagnose} failed atomic requirements, \emph{Allocate} a type-conditioned operator to each failure, \emph{Compile} the allocated local repairs into one globally coherent prompt, and \emph{Adopt} the repaired output only when it is a reliable replacement. The following subsections instantiate these four decisions without training or extra repaired-image candidates.

\subsection{Overview and Problem Setup}
We cast prompt optimization as single-pass visual feedback. Let $G$ be a frozen text-to-image generator and $p$ a user prompt.
Following DSG~\citep{cho2024dsg}, $p$ is decomposed into atomic semantic propositions $Q(p)=\{q_1,\dots,q_n\}$, each carrying a natural-language verification question and a semantic category.
A frozen VLM judge $V$ answers each question on an image $x$, and we define the semantic score
\begin{equation}
s(x) \;=\; \frac{1}{n}\,\bigl|\{\, i : V(x,q_i)=\textsc{correct} \,\}\bigr|,
\label{eq:score}
\end{equation}
computed with the same DSG evaluator and judge for every method, so all methods are scored under an identical protocol.\looseness=-1

TARA first renders $x_0=G(p)$ and obtains diagnosis $d_0$. Let
\begin{equation}
\begin{aligned}
F_0(x_0,p) &= \{(q_i,t_i):q_i\text{ is diagnosed as failed on }x_0\},\\
F &= F_0\cup\mathcal{R}(d_0),
& r_i &= \mathcal{A}(q_i,t_i),\\
\hat{p} &= \mathcal{C}\!\left(p,\{r_i\}_{(q_i,t_i)\in F}\right),
\end{aligned}
\label{eq:allocation}
\end{equation}
where $t_i$ is the diagnosed error type and $\mathcal{R}(d_0)$ recovers pruned relation/action constraints needed for repair. The allocator $\mathcal{A}$ maps each target to a local repair constraint $r_i$, and the compiler $\mathcal{C}$ realizes all allocated repairs through one prompt. In TARA, a repair operator is a fixed local language-transformation rule rather than a learned router. Uniform expansion is the type-invariant special case
\begin{equation}
\mathcal{A}_{\mathrm{uni}}(q,t)=R_{\mathrm{uni}}(q)\qquad\text{for every type }t,
\label{eq:uniform_special}
\end{equation}
so the resulting constraint may depend on the failed proposition $q$, but the repair policy does not change with its type $t$.

\begin{figure}[h!]
\centering
\includegraphics[width=0.8\textwidth]{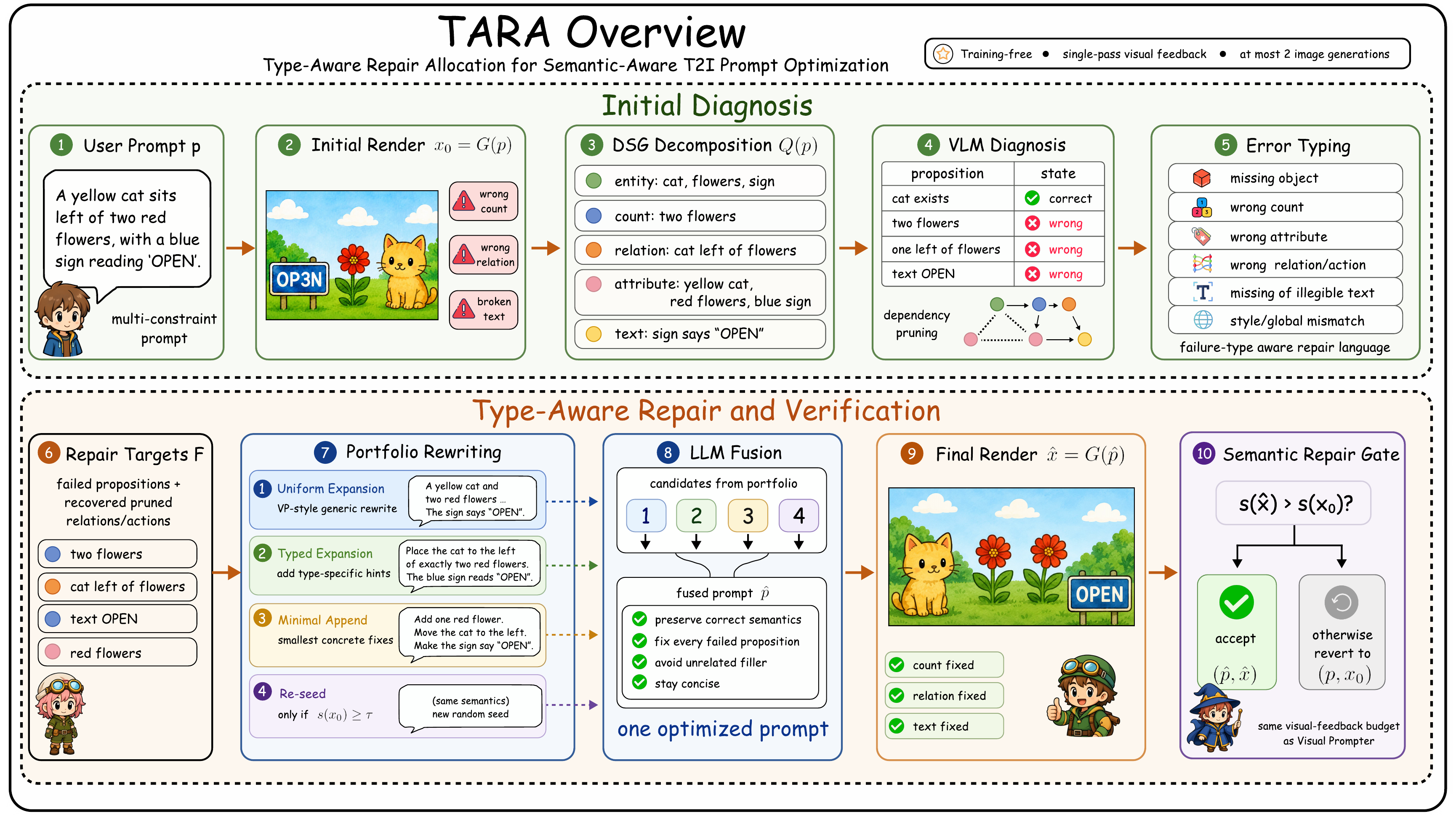}
\caption{Overview of TARA. It diagnoses failed atomic propositions, allocates type-conditioned repairs, compiles them into one prompt, and adopts the single repaired image only if its semantic score improves. Training-free, within at most two image generations.}
\label{fig:pipeline}
\end{figure}

Operationally, TARA (i) diagnoses the failed propositions in $x_0$; (ii) allocates each failure to a type-conditioned repair operator; (iii) compiles the resulting local constraints into one optimized prompt $\hat{p}$; and (iv) renders a single repaired image $\hat{x}=G(\hat{p})$, which is adopted only if it improves the score and is otherwise discarded in favor of $x_0$.
TARA thus uses at most two image generations, one diagnostic and one repaired, while all diagnosis, candidate construction, selection, and fusion are text- or VLM-side.
Figure~\ref{fig:pipeline} summarizes the two stages (initial diagnosis, then type-aware repair with a final gate), and Algorithm~\ref{alg:tara} in the appendix gives the full procedure.

\subsection{Typed Visual Diagnosis}
Following DSG~\citep{cho2024dsg}, each prompt is decomposed into atomic propositions, and the VLM judge assigns every proposition one of three states (\emph{correct}, \emph{absent} when not generated, or \emph{wrong} when generated but incorrect), together with a semantic category (entity, attribute, relation, action, count, text, style, or global).
The DSG dependency graph prunes invalid downstream questions: if an entity is absent, questions about its attributes or relations are not counted as independent failures.
Following the DSG evaluation protocol, a pruned proposition counts as zero in the numerator of Eq.~\ref{eq:score} but stays in its denominator, so TARA and all baselines are compared under an identical metric rather than a more lenient one.

TARA maps each failed proposition to a repairable error type through a fixed category-to-type table: entity $\rightarrow$ \emph{missing object}, count $\rightarrow$ \emph{wrong count}, attribute $\rightarrow$ \emph{wrong attribute}, relation $\rightarrow$ \emph{wrong relation}, action $\rightarrow$ \emph{wrong action}, text $\rightarrow$ \emph{missing or illegible text}, style $\rightarrow$ \emph{style mismatch}, and global $\rightarrow$ \emph{global mismatch}. Through this fixed mapping, shared across all benchmarks and generators, TARA organizes diagnosed failures into eight repairable types that act as actionable routing variables.
The type determines the repair language: explicit cardinality and separation for counts, unambiguous spatial anchors for relations, visible pose and contact for actions, an exact quoted string with legible typography for text, and salient foreground grounding for missing objects, all while leaving already-correct content untouched. These typed failures instantiate the base set $F_0$ in Eq.~\ref{eq:allocation}.\looseness=-1

A failed entity prunes its dependent relations, which can trap an optimizer into repeatedly re-adding the missing object while never re-describing the layout.
TARA therefore \emph{also} adds pruned relation and action propositions that are themselves wrong or absent into $F$, so that a single rewrite can re-describe a missing object together with its spatial relation.
This recovery affects only the repair targets, not the comparable score of Eq.~\ref{eq:score}.\looseness=-1

\subsection{Typed Repair Compilation and Output Adoption}\label{subsec:portfolio}
Once each failed proposition has been allocated a repair operator, TARA must compile heterogeneous local constraints into one globally coherent prompt. Directly committing to one rewrite bias is brittle: a conservative edit may under-specify a missing concept, while an aggressive one may inject unrelated detail. TARA therefore implements typed repair compilation with a small portfolio $C$ of text-only rewrite drafts, each encoding a different compilation bias:
\begin{enumerate}
    \item \textbf{Uniform expansion}: a type-agnostic rewrite that enriches $p$ so all failed aspects are depicted at once, with no type information---mirroring the uniform-expansion bias of prior optimizers and serving as an in-method control for typing. Like every candidate, it is a non-final draft built from the recovered targets $F$ and later passed to fusion.
    \item \textbf{Typed expansion}: the same expansion, but each failed aspect is annotated with its error type and a type-specific repair phrase (word-and-digit counts, spatial anchors, quoted legible text, pose and contact for actions).
    \item \textbf{Minimal append}: keeps $p$ almost verbatim and appends only the shortest concrete clauses needed to fix the failures, for strong generators where a full rewrite tends to drift from the original intent.
\end{enumerate}
When the initial image is already close to correct ($s(x_0)\ge\tau$, with $\tau{=}0.72$), TARA additionally admits a \textbf{re-seed} candidate that retains $p$ unchanged, allowing the single repair-generation step to use a fresh seed instead of a rewritten prompt---avoiding harmful edits to near-correct images. We use $\tau{=}0.72$ as a near-correct cutoff (roughly ``most propositions already pass''); a threshold sweep (Appendix~\ref{app:ablation_sensitivity}) shows that all values in $[0.60,0.84]$ give large gains at a comparable image budget, and we keep this single global $\tau$ fixed rather than tuning it per generator or benchmark.
Because every candidate is text-only, the portfolio adds no image generations.
The portfolio is not an image-level candidate set: none of its drafts is rendered or scored as an image. Instead, it provides complementary textual realizations of the allocated local repairs so they can be compiled into one prompt under the single-regeneration budget.

TARA then fuses the portfolio into one optimized prompt $\hat{p}$ with a text-only LLM step whose instruction enforces four principles: preserve all already-correct semantics, fix every failed proposition, avoid unrelated objects and generic aesthetic filler, and stay concise.
As an implementation alternative, we also consider a purely heuristic selector that instead picks one existing candidate by a text-only score rewarding coverage of failed-proposition words and preservation of the original wording while penalizing over-long prompts; it never inspects an image, preserving the single-pass budget.

Even a type-targeted rewrite may remove previously correct content. TARA therefore ends its inference policy with a semantic repair gate: after generating exactly one prescribed repair $\hat{x}=G(\hat{p})$, it adopts the repair iff $s(\hat{x})>s(x_0)$ and otherwise reverts to $(p,x_0)$. This accept-or-revert regression guard is not a best-of-$N$ selector: it generates no additional repaired candidates and performs no image search. It operates per input without ground-truth annotations or reference images, and its full cost is included in our runtime accounting (Appendix~\ref{app:implementation}).\looseness=-1

\section{Experiments}

We organize our study around four questions, answered in turn:
\begin{itemize}
    \item \textbf{Q1 (accuracy).} Does TARA produce more semantically accurate images than existing prompt optimizers?
    \item \textbf{Q2 (quality).} Does it do so without sacrificing CLIP alignment or aesthetics?
    \item \textbf{Q3 (allocation).} Does explicit type-conditioned repair contribute, and do gains hold across semantic categories and generators?
    \item \textbf{Q4 (reliability and efficiency).} Do repair compilation and adoption contribute under the matched budget, and is TARA efficient?
\end{itemize}
Implementation details, repair templates, seed robustness, proposition-type breakdowns, and preference numbers are in the Appendix.

\subsection{Experiment Setup}
We evaluate on DSG-1k~\citep{cho2024dsg} (10 source datasets) and TIFA v1.0~\citep{hu2023tifa}, running three seeds with 200 stratified prompts per seed for each benchmark-generator-method cell.\looseness=-1

Our primary metric is Semantic Accuracy: the fraction of atomic propositions answered ``yes'' by the shared VLM judge (pruned propositions count as zero but stay in the denominator). We also report CLIP Score~\citep{hessel2021clipscore}, Aesthetic Score~\citep{schuhmann2022laion}, and a VLM-as-Judge preference study. The VLM never sees the optimized prompt text, so a rewrite cannot cue the evaluator.\looseness=-1

Four generators: SD~v1.5, SD~v2.1~\citep{rombach2022high}, Flux-dev~\citep{blackforestlabs2024flux}, and Janus-Pro~\citep{chen2025januspro}. Baselines: Promptist~\citep{hao2023optimizing}, BeautifulPrompt~\citep{cao2023beautifulprompt}, NeuroPrompts~\citep{rosenman2024neuroprompts}, TIPO~\citep{yeh2024tipo}, and VisualPrompter~\citep{wu2026visualprompter}. All methods share prompts, seeds, image sizes, judge, and evaluation protocol, and each is evaluated in its native end-to-end configuration under the same maximum image-generation budget: VisualPrompter follows its published pipeline, whereas TARA includes the semantic repair gate as part of its inference procedure, with all additional model calls and end-to-end costs reported. External methods remain native; the gate's contribution is isolated by the controlled within-TARA ablation in \S\ref{subsec:a4}. The external VisualPrompter baseline is distinct from TARA's internal uniform-expansion candidate (\S\ref{subsec:portfolio}), a non-final draft used as an in-method control for typing. Details are in Appendix~\ref{app:implementation}.\looseness=-1

\subsection{A1: TARA Achieves the Best Semantic Accuracy in Every Cell}
TARA is the strongest method in all eight benchmark-generator cells. As shown in Tables~\ref{tab:persource_dsg} and~\ref{tab:persource_tifa} (whose Avg columns give the per-generator overall), TARA improves over VisualPrompter by \textbf{+5.6} points on DSG (76.4$\pm$0.2 vs.\ 70.8$\pm$0.7) and \textbf{+2.6} on TIFA (85.4$\pm$0.2 vs.\ 82.8$\pm$0.6), with prompt-clustered paired bootstrap 95\% CIs of $[4.7,6.5]$ and $[1.6,3.6]$, respectively. Across 4{,}800 paired evaluations, the mean paired gain is $+4.11$ points (95\% CI $[3.45,4.80]$). Re-scoring the saved final images with four held-out VLM evaluators, none of which participates in diagnosis or gate decisions, confirms the ranking, with Overall gains ranging from $+2.4$ to $+3.2$ points (Table~\ref{tab:judge_robust}). In contrast, aesthetics-oriented optimizers such as BeautifulPrompt and TIPO fall well below the raw-prompt baseline, showing that generic enrichment can actively hurt semantics.\looseness=-1


\begin{wraptable}{r}{0.58\textwidth}
\vspace{-0.3em}
\centering
\caption{Evaluator robustness for VP and TARA. Scores are averages over four generators.}
\label{tab:judge_robust}
\scriptsize
\setlength{\tabcolsep}{2.2pt}
\resizebox{\linewidth}{!}{%
\begin{tabular}{l ccc ccc ccc}
\toprule
\multirow{2}{*}{Evaluator} & \multicolumn{3}{c}{DSG} & \multicolumn{3}{c}{TIFA} & \multicolumn{3}{c}{Overall} \\
\cmidrule(lr){2-4}\cmidrule(lr){5-7}\cmidrule(lr){8-10}
& VP & TARA & Gap & VP & TARA & Gap & VP & TARA & Gap \\
\midrule
Qwen3.5-4B & 72.9 & \textbf{76.1} & \cellcolor{cPos!32}+3.2 & 82.6 & \textbf{84.6} & \cellcolor{cPos!20}+2.0 & 77.7 & \textbf{80.4} & \cellcolor{cPos!26}+2.6 \\
\shortstack[l]{Qwen2.5-VL-\\7B-Instruct} & 71.0 & \textbf{74.0} & \cellcolor{cPos!30}+3.0 & 80.5 & \textbf{82.4} & \cellcolor{cPos!19}+1.9 & 75.8 & \textbf{78.2} & \cellcolor{cPos!24}+2.4 \\
Gemma4-31B & 69.7 & \textbf{72.4} & \cellcolor{cPos!27}+2.7 & 79.2 & \textbf{81.3} & \cellcolor{cPos!21}+2.1 & 74.5 & \textbf{76.9} & \cellcolor{cPos!24}+2.4 \\
Qwen3.5-27B & 71.2 & \textbf{75.3} & \cellcolor{cPos!41}+4.1 & 82.6 & \textbf{84.8} & \cellcolor{cPos!22}+2.2 & 76.9 & \textbf{80.1} & \cellcolor{cPos!32}+3.2 \\
\bottomrule
\end{tabular}}
\vspace{-0.4em}
\end{wraptable}

Because visual feedback methods can be sensitive to the VLM used for diagnosis and scoring, we also test evaluator robustness.
Keeping the generated images fixed, we re-run only the atomic yes/no evaluation for the strongest semantic baseline, VisualPrompter, and TARA using four held-out VLM judges.
Table~\ref{tab:judge_robust} shows that the ranking does not depend on the main evaluator: TARA remains better on both DSG and TIFA averages under all four judges.

The gains are not uniform across difficulty: TARA improves the most on DSG, whose prompts carry more explicit counts, relations, text, and unusual compositions, exactly the cases where a single uniform expansion is least adequate and type-aware repair helps most.

\begin{table}[htbp]
\centering
\scriptsize
\caption{Per-source semantic accuracy (\%) on DSG across four generators and seven methods (3-seed mean$\pm$std; per-column best in bold; colored row is TARA$-$VisualPrompter).}
\label{tab:persource_dsg}
\setlength{\tabcolsep}{3pt}\renewcommand{\arraystretch}{1.1}
\resizebox{\textwidth}{!}{%
\begin{tabular}{l l c c c c c c c c c c c }
\toprule
Generator & Method & Whoops & Localized & PoseScript & VRD & CountBench & Midjourney & TIFA160 & DrawText & Stanford & DiffusionDB & Avg \\
\midrule
SD 1.5 & Raw prompt & 74.0\std{1.4} & 69.1\std{3.1} & 60.3\std{1.6} & 60.4\std{4.1} & 51.2\std{2.6} & 57.5\std{3.6} & 67.9\std{5.5} & 53.2\std{3.4} & 69.5\std{2.6} & 53.6\std{4.5} & 61.7\std{0.5} \\
 & NeuroPrompts & 63.2\std{2.1} & 56.2\std{2.9} & 48.9\std{4.7} & 63.9\std{2.8} & 38.2\std{4.4} & 38.9\std{3.0} & 45.6\std{6.7} & 37.1\std{3.9} & 62.6\std{3.5} & 47.4\std{1.6} & 50.2\std{1.5} \\
 & Promptist & 70.8\std{6.6} & 64.0\std{0.7} & 48.3\std{3.3} & 65.1\std{2.9} & 42.1\std{5.9} & 50.4\std{3.1} & 55.3\std{2.8} & 42.5\std{1.2} & 62.0\std{0.6} & 49.1\std{0.7} & 55.0\std{1.3} \\
 & BeautifulPrompt & 52.9\std{7.2} & 42.9\std{4.8} & 36.7\std{2.0} & 49.6\std{5.5} & 35.4\std{6.6} & 31.4\std{4.3} & 38.8\std{2.8} & 30.7\std{5.6} & 31.0\std{4.8} & 33.1\std{6.5} & 38.2\std{0.8} \\
 & TIPO & 59.4\std{3.3} & 58.4\std{2.2} & 56.1\std{7.3} & 62.8\std{3.7} & 40.5\std{3.7} & 50.2\std{0.9} & 57.6\std{5.5} & 47.1\std{0.7} & 62.6\std{2.8} & 47.1\std{2.4} & 54.2\std{1.1} \\
\rowcolor{cSecond}  & VisualPrompter & 77.7\std{5.1} & 70.4\std{2.2} & 57.9\std{1.3} & 67.0\std{2.9} & 61.1\std{2.1} & 53.8\std{3.8} & 69.1\std{2.0} & 48.9\std{2.9} & 63.5\std{4.1} & 57.3\std{2.4} & 62.7\std{1.3} \\
\rowcolor{cBest}  & TARA (Ours) & \textbf{80.4}\std{4.4} & \textbf{74.2}\std{0.9} & \textbf{66.3}\std{2.8} & \textbf{77.9}\std{2.8} & \textbf{62.7}\std{6.1} & \textbf{65.2}\std{2.2} & \textbf{79.1}\std{1.5} & \textbf{60.9}\std{1.0} & \textbf{74.5}\std{2.5} & \textbf{57.6}\std{5.0} & \textbf{69.9}\std{0.7} \\
 & {\itshape TARA$-$VP} & \cellcolor{cPos!26}+2.7 & \cellcolor{cPos!36}+3.8 & \cellcolor{cPos!80}+8.4 & \cellcolor{cPos!80}+10.9 & \cellcolor{cPos!15}+1.6 & \cellcolor{cPos!80}+11.4 & \cellcolor{cPos!80}+10.0 & \cellcolor{cPos!80}+12.1 & \cellcolor{cPos!80}+11.0 & \cellcolor{cPos!8}+0.4 & \cellcolor{cPos!69}+7.2 \\
\midrule
SD 2.1 & Raw prompt & 80.9\std{1.5} & 67.7\std{2.0} & 61.6\std{1.8} & 70.6\std{2.1} & 57.2\std{0.3} & 56.7\std{3.8} & 70.5\std{2.0} & 57.7\std{3.2} & 75.9\std{3.3} & 52.4\std{4.7} & 65.1\std{1.5} \\
 & NeuroPrompts & 58.4\std{3.2} & 69.9\std{1.6} & 52.2\std{2.3} & 68.0\std{1.9} & 53.7\std{2.0} & 49.7\std{2.1} & 69.8\std{2.8} & 49.0\std{3.8} & 69.6\std{2.8} & 50.1\std{0.8} & 59.0\std{0.3} \\
 & Promptist & 69.3\std{2.9} & 73.5\std{2.9} & 56.5\std{1.4} & 66.4\std{7.4} & 53.0\std{3.7} & 53.2\std{1.7} & 69.1\std{3.0} & 50.4\std{4.7} & 69.2\std{1.4} & 52.1\std{3.4} & 61.3\std{1.1} \\
 & BeautifulPrompt & 50.8\std{2.8} & 43.3\std{8.7} & 41.2\std{6.1} & 51.1\std{1.1} & 33.1\std{1.1} & 33.0\std{4.8} & 44.5\std{7.6} & 35.3\std{5.5} & 31.8\std{1.8} & 36.4\std{9.4} & 40.1\std{1.0} \\
 & TIPO & 51.9\std{3.4} & 61.1\std{3.1} & 56.7\std{0.5} & 59.7\std{5.5} & 43.0\std{6.3} & 39.5\std{3.6} & 57.8\std{2.5} & 43.6\std{1.6} & 66.3\std{3.2} & 42.7\std{3.4} & 52.2\std{0.8} \\
\rowcolor{cSecond}  & VisualPrompter & 79.8\std{2.1} & 71.2\std{4.0} & 63.9\std{0.3} & 77.3\std{2.3} & 53.9\std{3.7} & 58.7\std{4.3} & 73.5\std{3.7} & 55.7\std{2.5} & 77.7\std{2.5} & 54.5\std{2.2} & 66.6\std{0.8} \\
\rowcolor{cBest}  & TARA (Ours) & \textbf{87.1}\std{1.1} & \textbf{77.3}\std{2.5} & \textbf{68.6}\std{0.8} & \textbf{82.3}\std{1.1} & \textbf{65.3}\std{1.3} & \textbf{65.5}\std{5.6} & \textbf{79.0}\std{1.3} & \textbf{60.4}\std{0.8} & \textbf{84.5}\std{4.4} & \textbf{62.6}\std{2.3} & \textbf{73.2}\std{0.5} \\
 & {\itshape TARA$-$VP} & \cellcolor{cPos!70}+7.3 & \cellcolor{cPos!58}+6.1 & \cellcolor{cPos!45}+4.7 & \cellcolor{cPos!48}+5.0 & \cellcolor{cPos!80}+11.3 & \cellcolor{cPos!66}+6.8 & \cellcolor{cPos!53}+5.5 & \cellcolor{cPos!45}+4.7 & \cellcolor{cPos!66}+6.8 & \cellcolor{cPos!77}+8.0 & \cellcolor{cPos!64}+6.6 \\
\midrule
Flux-dev & Raw prompt & 82.4\std{3.1} & 85.7\std{2.3} & 73.2\std{4.1} & 85.5\std{1.2} & 67.7\std{1.7} & 46.8\std{2.7} & 79.9\std{1.8} & 75.0\std{3.6} & 92.8\std{1.2} & 56.2\std{2.7} & 74.5\std{1.0} \\
 & NeuroPrompts & 76.7\std{3.0} & 76.6\std{2.8} & 67.3\std{6.3} & 87.6\std{2.7} & 60.7\std{2.7} & 41.7\std{2.7} & 83.8\std{1.1} & 64.8\std{0.7} & 89.5\std{2.2} & 49.2\std{1.6} & 69.8\std{0.6} \\
 & Promptist & 78.6\std{7.0} & 84.4\std{2.8} & 66.4\std{1.3} & 84.1\std{2.2} & 67.0\std{5.9} & 48.2\std{0.9} & 79.3\std{1.1} & 69.8\std{1.4} & 86.6\std{2.2} & 51.0\std{2.8} & 71.5\std{0.8} \\
 & BeautifulPrompt & 60.5\std{3.3} & 44.8\std{7.7} & 49.1\std{2.3} & 63.8\std{4.9} & 36.7\std{4.2} & 29.7\std{4.1} & 49.0\std{7.7} & 35.6\std{1.2} & 37.0\std{1.8} & 43.1\std{6.5} & 44.9\std{1.0} \\
 & TIPO & 73.3\std{3.5} & 77.1\std{1.6} & 65.3\std{2.7} & 80.2\std{3.4} & 54.9\std{2.9} & 34.8\std{4.6} & 72.6\std{5.9} & 64.8\std{3.5} & 87.2\std{2.8} & 44.9\std{3.0} & 65.5\std{1.5} \\
\rowcolor{cSecond}  & VisualPrompter & 85.7\std{3.6} & 85.3\std{2.5} & 72.3\std{3.0} & 92.2\std{2.4} & 69.7\std{3.3} & 53.4\std{3.4} & 88.3\std{0.3} & 83.1\std{0.6} & 92.7\std{1.9} & \textbf{62.9}\std{0.9} & 78.6\std{1.1} \\
\rowcolor{cBest}  & TARA (Ours) & \textbf{88.4}\std{1.3} & \textbf{87.8}\std{1.6} & \textbf{78.5}\std{1.7} & \textbf{95.8}\std{0.8} & \textbf{77.6}\std{1.4} & \textbf{58.1}\std{1.9} & \textbf{89.4}\std{1.5} & \textbf{86.3}\std{2.4} & \textbf{94.7}\std{0.6} & 61.6\std{1.2} & \textbf{81.8}\std{0.5} \\
 & {\itshape TARA$-$VP} & \cellcolor{cPos!26}+2.7 & \cellcolor{cPos!25}+2.6 & \cellcolor{cPos!60}+6.2 & \cellcolor{cPos!34}+3.5 & \cellcolor{cPos!75}+7.8 & \cellcolor{cPos!45}+4.7 & \cellcolor{cPos!11}+1.2 & \cellcolor{cPos!31}+3.3 & \cellcolor{cPos!20}+2.1 & \cellcolor{cNeg!26}$-$1.3 & \cellcolor{cPos!31}+3.3 \\
\midrule
Janus-Pro & Raw prompt & 80.4\std{1.6} & 82.1\std{1.8} & 61.0\std{3.7} & 83.7\std{1.5} & 56.0\std{7.5} & 52.8\std{4.5} & 81.1\std{1.5} & 75.8\std{5.6} & 85.5\std{1.6} & 58.7\std{5.2} & 71.7\std{1.3} \\
 & NeuroPrompts & 80.8\std{2.5} & 78.5\std{1.8} & 61.6\std{3.8} & 85.8\std{0.8} & 57.6\std{2.4} & 47.3\std{2.7} & 84.0\std{2.4} & 69.3\std{1.9} & 87.6\std{2.8} & 53.7\std{0.8} & 70.6\std{0.6} \\
 & Promptist & 79.2\std{3.4} & 78.8\std{1.8} & 62.0\std{2.2} & 85.1\std{2.0} & 59.8\std{2.7} & 52.6\std{4.7} & 83.2\std{1.9} & 69.7\std{2.0} & 82.5\std{1.6} & 54.3\std{3.3} & 70.7\std{1.3} \\
 & BeautifulPrompt & 61.0\std{4.6} & 42.8\std{5.0} & 47.9\std{0.8} & 63.5\std{0.5} & 32.5\std{3.3} & 33.5\std{5.7} & 54.4\std{4.5} & 44.6\std{2.1} & 35.9\std{3.0} & 39.9\std{8.0} & 45.6\std{1.9} \\
 & TIPO & 76.0\std{5.2} & 79.4\std{1.7} & 60.8\std{2.3} & 79.0\std{3.3} & 51.0\std{8.8} & 36.7\std{4.1} & 69.5\std{1.3} & 56.7\std{4.9} & 82.6\std{1.9} & 43.6\std{3.8} & 63.5\std{2.4} \\
\rowcolor{cSecond}  & VisualPrompter & 85.1\std{2.0} & 82.9\std{3.4} & 68.0\std{1.4} & 89.7\std{1.3} & 64.1\std{3.8} & 54.3\std{1.9} & 83.4\std{1.9} & 76.5\std{1.5} & 90.1\std{1.5} & 58.4\std{2.6} & 75.2\std{0.4} \\
\rowcolor{cBest}  & TARA (Ours) & \textbf{89.5}\std{0.6} & \textbf{85.5}\std{0.7} & \textbf{68.4}\std{2.8} & \textbf{91.7}\std{1.4} & \textbf{73.2}\std{3.7} & \textbf{64.9}\std{1.9} & \textbf{88.0}\std{2.3} & \textbf{84.0}\std{4.2} & \textbf{93.5}\std{1.3} & \textbf{66.9}\std{4.0} & \textbf{80.6}\std{1.1} \\
 & {\itshape TARA$-$VP} & \cellcolor{cPos!43}+4.5 & \cellcolor{cPos!25}+2.6 & \cellcolor{cPos!8}+0.4 & \cellcolor{cPos!19}+2.0 & \cellcolor{cPos!80}+9.1 & \cellcolor{cPos!80}+10.7 & \cellcolor{cPos!44}+4.6 & \cellcolor{cPos!72}+7.5 & \cellcolor{cPos!32}+3.3 & \cellcolor{cPos!80}+8.5 & \cellcolor{cPos!51}+5.3 \\
\bottomrule
\end{tabular}}
\end{table}

\begin{table}[t]
\centering
\scriptsize
\caption{Per-source semantic accuracy (\%) on TIFA across four generators and seven methods (transposed: rows are sources, columns are methods). Conventions as in Table~\ref{tab:persource_dsg}.}
\label{tab:persource_tifa}
\setlength{\tabcolsep}{3pt}\renewcommand{\arraystretch}{1.1}
\resizebox{\textwidth}{!}{%
\begin{tabular}{l l c c c c c c c c}
\toprule
Generator & Source & Raw & Neuro & Promptist & Beauty & TIPO & VP & TARA & TARA$-$VP \\
\midrule
SD 1.5 & COCO & 75.6\std{2.8} & 61.3\std{2.6} & 71.9\std{0.2} & 51.9\std{2.6} & 63.4\std{2.2} & 82.8\std{4.3} & \textbf{85.3}\std{0.4} & \cellcolor{cPos!24}+2.5 \\
 & DrawBench & 56.6\std{2.8} & 44.0\std{4.1} & 54.6\std{3.0} & 40.5\std{3.6} & 35.3\std{5.3} & 62.0\std{1.9} & \textbf{66.1}\std{2.1} & \cellcolor{cPos!40}+4.2 \\
 & PartiPrompt & 72.1\std{2.9} & 48.1\std{4.4} & 71.0\std{2.9} & 41.3\std{4.4} & 49.3\std{2.8} & 79.0\std{2.4} & \textbf{80.9}\std{0.6} & \cellcolor{cPos!18}+1.9 \\
 & PaintSkill & 59.4\std{1.3} & 41.8\std{1.0} & 54.8\std{1.5} & 37.6\std{2.0} & 38.6\std{7.1} & 65.6\std{3.6} & \textbf{70.8}\std{3.3} & \cellcolor{cPos!50}+5.2 \\
 & Avg & 66.0\std{0.5} & 48.8\std{0.8} & 63.0\std{0.8} & 42.8\std{0.3} & 46.6\std{1.2} & 72.3\std{2.5} & \textbf{75.8}\std{1.4} & \cellcolor{cPos!33}+3.5 \\
\midrule
SD 2.1 & COCO & 80.8\std{2.3} & 73.8\std{2.0} & 78.0\std{3.6} & 56.3\std{0.9} & 65.3\std{0.9} & 86.2\std{2.6} & \textbf{88.8}\std{0.8} & \cellcolor{cPos!25}+2.6 \\
 & DrawBench & 64.5\std{0.9} & 57.7\std{1.7} & 64.1\std{0.9} & 44.4\std{2.5} & 38.0\std{0.1} & 66.4\std{1.6} & \textbf{73.6}\std{1.7} & \cellcolor{cPos!69}+7.2 \\
 & PartiPrompt & 80.9\std{2.8} & 65.8\std{4.9} & 71.8\std{4.9} & 40.7\std{4.6} & 51.2\std{1.6} & 83.9\std{0.5} & \textbf{87.5}\std{1.9} & \cellcolor{cPos!34}+3.6 \\
 & PaintSkill & 68.3\std{0.0} & 61.2\std{3.4} & 67.2\std{2.2} & 45.8\std{2.8} & 47.2\std{4.7} & 73.9\std{1.2} & \textbf{77.3}\std{3.9} & \cellcolor{cPos!33}+3.4 \\
 & Avg & 73.6\std{0.1} & 64.6\std{2.0} & 70.3\std{0.2} & 46.8\std{1.2} & 50.4\std{0.6} & 77.6\std{0.1} & \textbf{81.8}\std{1.5} & \cellcolor{cPos!40}+4.2 \\
\midrule
Flux-dev & COCO & 87.8\std{1.2} & 84.6\std{2.6} & 87.0\std{1.3} & 62.3\std{3.3} & 80.0\std{3.2} & 91.7\std{0.9} & \textbf{93.2}\std{0.9} & \cellcolor{cPos!14}+1.5 \\
 & DrawBench & 80.2\std{1.3} & 72.0\std{2.3} & 71.4\std{1.6} & 51.6\std{4.8} & 63.7\std{2.5} & 85.5\std{0.2} & \textbf{87.9}\std{1.0} & \cellcolor{cPos!23}+2.4 \\
 & PartiPrompt & 93.2\std{0.7} & 84.9\std{0.6} & 89.1\std{1.8} & 46.9\std{5.0} & 77.9\std{6.0} & 93.8\std{1.2} & \textbf{94.9}\std{2.1} & \cellcolor{cPos!11}+1.1 \\
 & PaintSkill & 76.8\std{1.3} & 74.3\std{2.9} & 75.9\std{1.2} & 54.3\std{3.5} & 78.1\std{1.3} & 94.4\std{1.5} & \textbf{95.9}\std{2.8} & \cellcolor{cPos!14}+1.4 \\
 & Avg & 84.5\std{0.3} & 79.0\std{1.1} & 80.8\std{0.3} & 53.8\std{1.5} & 74.9\std{1.3} & 91.4\std{0.6} & \textbf{93.0}\std{1.6} & \cellcolor{cPos!15}+1.6 \\
\midrule
Janus-Pro & COCO & 89.6\std{0.8} & 90.1\std{1.6} & 90.1\std{1.3} & 63.0\std{2.0} & 80.6\std{2.7} & 93.0\std{2.4} & \textbf{94.7}\std{0.4} & \cellcolor{cPos!17}+1.7 \\
 & DrawBench & 75.2\std{0.8} & 80.9\std{1.7} & 73.1\std{2.4} & 55.0\std{2.9} & 60.8\std{1.5} & 82.6\std{0.4} & \textbf{86.9}\std{1.3} & \cellcolor{cPos!41}+4.3 \\
 & PartiPrompt & 77.9\std{7.2} & 78.7\std{1.0} & 83.4\std{3.0} & 44.6\std{3.5} & 69.1\std{3.6} & \textbf{90.0}\std{0.8} & 89.8\std{1.1} & \cellcolor{cNeg!15}$-$0.2 \\
 & PaintSkill & 80.8\std{3.1} & 84.3\std{0.9} & 82.4\std{2.7} & 56.6\std{4.2} & 72.0\std{2.5} & \textbf{93.1}\std{1.7} & 92.1\std{1.5} & \cellcolor{cNeg!20}$-$1.0 \\
 & Avg & 80.9\std{0.7} & 83.5\std{0.8} & 82.3\std{1.4} & 54.8\std{1.4} & 70.6\std{0.9} & 89.7\std{0.8} & \textbf{90.9}\std{0.4} & \cellcolor{cPos!12}+1.2 \\
\bottomrule
\end{tabular}}
\end{table}

Tables~\ref{tab:persource_dsg} and~\ref{tab:persource_tifa} break down DSG and TIFA by source category across all four generators.
The largest gains appear in sources that emphasize difficult semantic structure: CountBench requires explicit cardinality, PoseScript and VRD require relations and actions, and DrawText requires robust text constraints.
This supports the central claim that atomic repair allocation is more effective than absorbing heterogeneous failures into a single uniform expansion.

\begin{figure}[h!]
\centering
\includegraphics[width=0.88\textwidth]{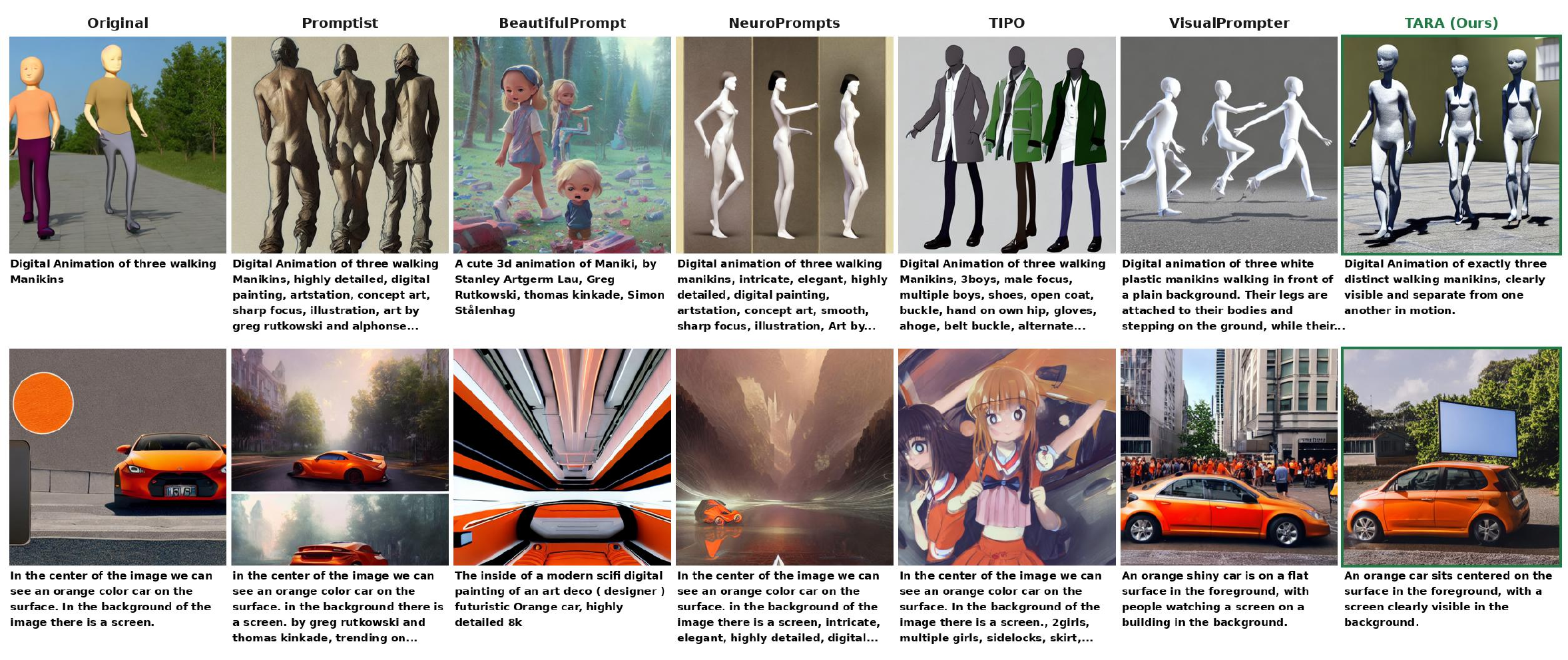}
\caption{Qualitative comparison on SD~v1.5. TARA (green) fixes the count (top) and the missing screen (bottom) that baselines miss.}
\label{fig:qualitative}
\end{figure}

Figure~\ref{fig:qualitative} compares all seven methods on two SD~v1.5 cases. In the count case (top), most baselines render the wrong number of manikins and aesthetics-oriented methods drift to stylized figures, whereas TARA yields three clean, countable manikins. In the object/relation case (bottom), only TARA keeps both the orange car and the background screen. Additional cases are in Figure~\ref{fig:qual_appendix}.\looseness=-1

\subsection{A2: TARA Preserves Image Quality While Fixing Semantics}
Higher semantic accuracy does not cost image quality. From Table~\ref{tab:clip} and Table~\ref{tab:aes} on DSG, we observe that
\begin{enumerate}
    \item TARA attains the highest average CLIP score and the best CLIP score on all four generators (mean $31.86$ vs.\ $31.45$ for VisualPrompter), so its prompts stay tightly aligned with the rendered images.
    \item Aesthetic-oriented baselines such as NeuroPrompts and BeautifulPrompt reach higher aesthetic scores but suffer large semantic drops, exposing a quality-for-semantics trade-off that TARA avoids.
    \item TARA stays within $0.02$ of VisualPrompter and the raw prompt in aesthetic score while improving semantic alignment, adding correctness without degrading visual appeal. This is a direct consequence of typed repair compilation: by targeting only the diagnosed failures with short, specific clauses rather than injecting generic aesthetic filler, TARA avoids the style drift that afflicts verbose expansion methods.
\end{enumerate}

\begin{table}[h!]
\begin{minipage}[t]{0.48\textwidth}
\centering
\caption{CLIP Score on DSG.}
\label{tab:clip}
\resizebox{\textwidth}{!}{%
\begin{tabular}{l cccc c}
\toprule
Method & SD1.5 & SD2.1 & Flux & Janus & Mean \\
\midrule
Raw prompt & 31.66 & 31.67 & 31.53 & 31.74 & 31.65 \\
NeuroPrompts & 28.91 & 29.86 & 30.63 & 31.29 & 30.17 \\
Promptist & 30.43 & 30.79 & 31.22 & 31.20 & 30.91 \\
BeautifulPrompt & 25.89 & 26.11 & 26.11 & 26.94 & 26.26 \\
TIPO & 28.37 & 27.32 & 27.68 & 27.77 & 27.78 \\
\midrule
\rowcolor{cSecond} VisualPrompter & 30.86 & 31.59 & 31.61 & 31.74 & 31.45 \\
\rowcolor{cBest} TARA (Ours) & \textbf{31.71} & \textbf{31.96} & \textbf{31.71} & \textbf{32.05} & \textbf{31.86} \\
\bottomrule
\end{tabular}}
\end{minipage}
\hfill
\begin{minipage}[t]{0.48\textwidth}
\centering
\caption{Aesthetic Score on DSG.}
\label{tab:aes}
\resizebox{\textwidth}{!}{%
\begin{tabular}{l cccc c}
\toprule
Method & SD1.5 & SD2.1 & Flux & Janus & Mean \\
\midrule
Raw prompt & 5.30 & 5.40 & 5.87 & 5.46 & 5.51 \\
\rowcolor{cBest} NeuroPrompts & \textbf{6.13} & 6.02 & 6.51 & 6.26 & \textbf{6.23} \\
Promptist & 5.99 & 5.89 & 6.35 & 6.01 & 6.06 \\
\rowcolor{cSecond} BeautifulPrompt & 6.01 & 5.98 & \textbf{6.53} & \textbf{6.27} & 6.20 \\
TIPO & 5.33 & 5.54 & 6.02 & 5.73 & 5.66 \\
\midrule
VisualPrompter & 5.31 & 5.42 & 5.72 & 5.45 & 5.48 \\
TARA (Ours) & 5.31 & 5.42 & 5.80 & 5.48 & 5.50 \\
\bottomrule
\end{tabular}}
\end{minipage}
\end{table}

\subsection{A3: TARA Performs Consistently across Types and Generators}
\begin{figure}[h!]
\centering
\includegraphics[width=0.9\textwidth]{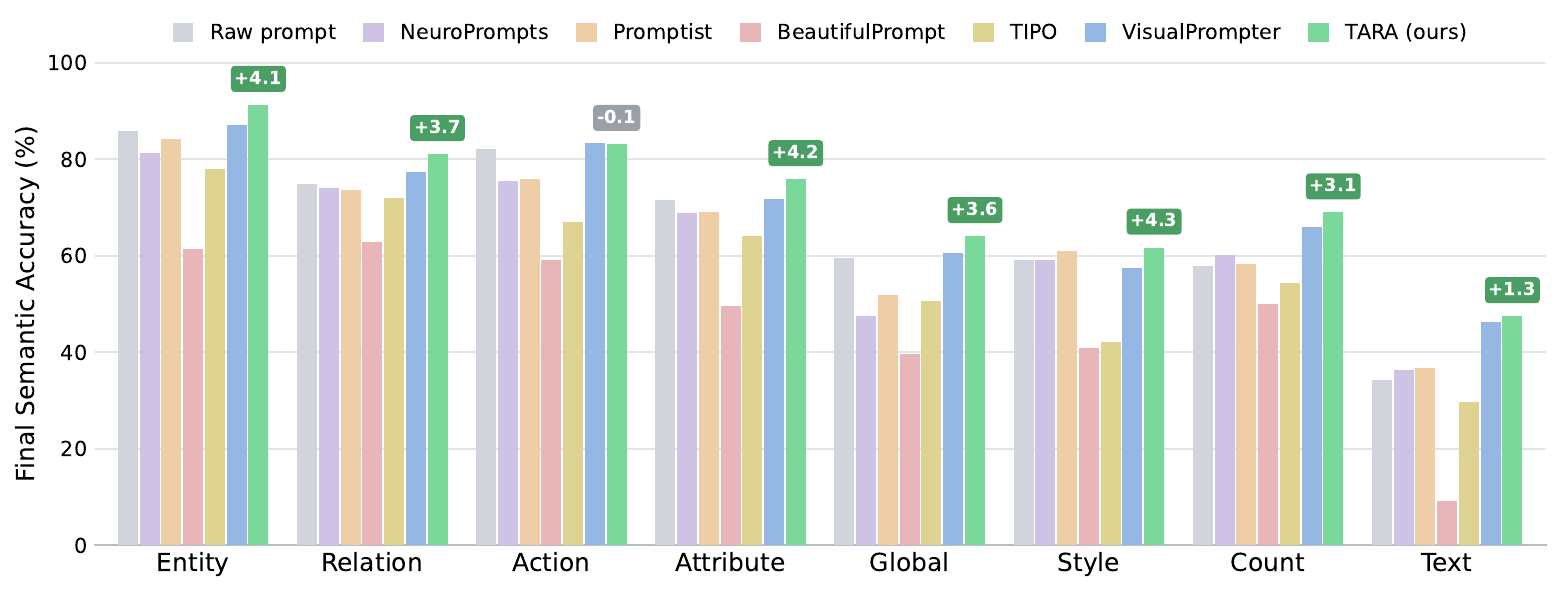}
\caption{Per-type semantic accuracy of seven methods over eight categories, aggregated over 24 benchmark--generator--seed cells; the pill above each green bar is TARA's gain over VisualPrompter.}
\label{fig:pertype}
\end{figure}

The improvement is broad, not a trade between failure types. Aggregating over all 24 benchmark--generator--seed cells (Figure~\ref{fig:pertype}), TARA improves over the raw prompt on all eight semantic categories and over \emph{every} prompt-optimization baseline on seven of them, essentially matching VisualPrompter on action ($83.2$ vs.\ $83.3$), rather than improving one category at another's expense. Action and style are the two rarest categories, so their per-method margins carry wider sampling noise; per-type repair behavior for all eight types is detailed in Table~\ref{tab:mech_errortype}. The gains are largest on style ($+4.3$), attribute ($+4.2$), entity ($+4.1$), and relation ($+3.7$)---exactly the types where uniform expansion under-specifies or drifts. A complementary VLM-as-Judge preference study (Appendix~\ref{app:pref}) favors TARA in $69.0\%$ of changed-prompt cases for semantic consistency. This per-type consistency is the empirical signature of type-aware repair: each failed proposition is rewritten with type-specific language, so gains accrue across categories rather than trading one for another.\looseness=-1

\begin{wrapfigure}{r}{0.50\textwidth}
\centering
\vspace{-0.5\baselineskip}
\includegraphics[width=\linewidth]{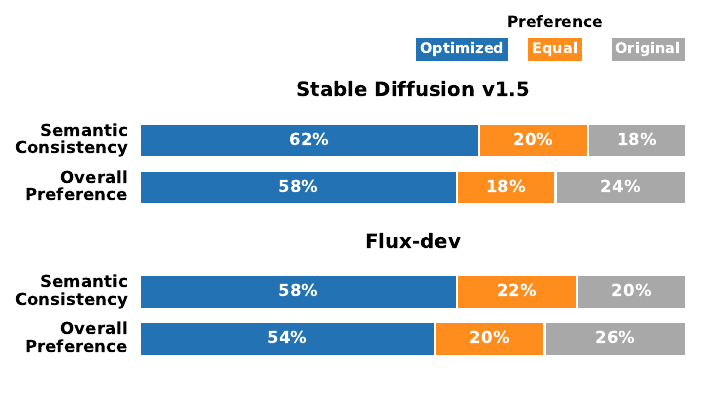}
\caption{Human preferences on 100 sampled cases under VisualPrompter's three-way protocol: TARA / Tie / Original.}
\label{fig:human_eval}
\vspace{-0.6\baselineskip}
\end{wrapfigure}
Following the human-evaluation protocol of VisualPrompter~\citep{wu2026visualprompter}, we compare TARA-optimized outputs against their original-prompt counterparts on 100 sampled cases from two representative generators, allowing raters to express equal preference. Human judgments corroborate the automatic results (Figure~\ref{fig:human_eval}). For semantic consistency, raters prefer TARA in 62\% of Stable Diffusion~v1.5 cases and 58\% of Flux-dev cases, compared with 18\% and 20\% for the original outputs. TARA is also favored in overall preference (58\% and 54\%, versus 24\% and 26\%), showing that its semantic gains translate to perceptually preferable images rather than metric-specific improvements.

TARA's typed prompts also show qualitative transfer to closed commercial systems; Appendix~\ref{app:online} gives examples on Doubao and GPT Image where TARA realizes the requested layout while baselines leave text incomplete or split objects.\looseness=-1

\subsection{A4: TARA Runs Faster than VisualPrompter, and Each Component Helps}\label{subsec:a4}
\begin{wraptable}{r}{0.48\textwidth}
\centering
\vspace{-0.3\baselineskip}
\caption{End-to-end time per prompt (s).}
\label{tab:time}
\resizebox{0.48\textwidth}{!}{%
\begin{tabular}{l ccccc}
\toprule
Method & SD1.5 & SD2.1 & Flux & Janus & Mean \\
\midrule
Raw prompt & 2.5 & 4.8 & 26.2 & 21.3 & 13.7 \\
NeuroPrompts & 11.6 & 14.5 & 50.7 & 35.6 & 28.1 \\
Promptist & 5.1 & 10.7 & 53.5 & 42.7 & 28.0 \\
BeautifulPrompt & 5.2 & 10.9 & 43.4 & 43.0 & 25.6 \\
TIPO & 11.1 & 14.2 & 41.6 & 34.4 & 25.3 \\
\midrule
\rowcolor{cSecond} VisualPrompter & 9.5 & 12.5 & 31.3 & 26.7 & 20.0 \\
\rowcolor{cBest} TARA (Ours) & \textbf{5.4} & \textbf{7.7} & \textbf{27.9} & \textbf{23.2} & \textbf{16.0} \\
\bottomrule
\end{tabular}}
\end{wraptable}
TARA is the fastest optimizer in our setting. At a matched $1.64$ image generations per prompt, it averages $16.0$s versus $20.0$s for VisualPrompter and $25$--$28$s for the others (Table~\ref{tab:time}). Although TARA uses more cheap text-only calls, its repaired prompts are shorter (22.7 vs.\ 41.6 words) and repair is faster (19.37 vs.\ 25.47s). Repairs are attempted on roughly two-thirds of prompts (those whose initial diagnosis reports a failure), unrepaired prompts finish in about $10$s, and per-type adoption rates range from $22.5\%$ to $29.7\%$ (Table~\ref{tab:mech_errortype}). Short, targeted rewrites thus improve both speed and accuracy; details are in Appendix~\ref{app:implementation}.\looseness=-1

\begin{wrapfigure}{r}{0.48\textwidth}
\centering
\vspace{-0.35\baselineskip}
\includegraphics[width=0.46\textwidth]{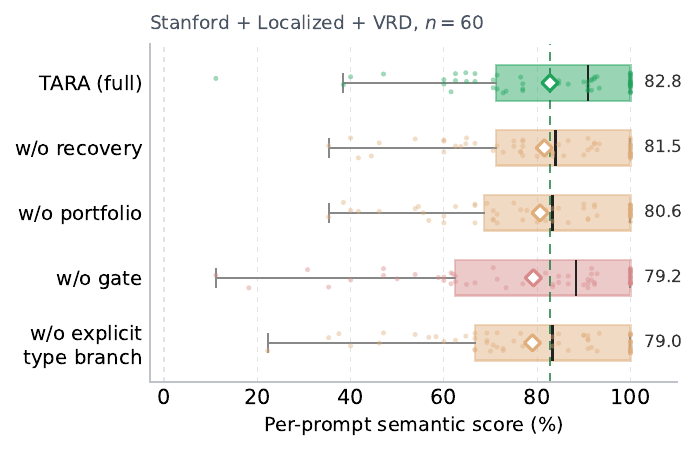}
\vspace{-0.6\baselineskip}
\caption{Full-scale DSG ablation across four generators and three seeds; diamonds denote means.}
\label{fig:ablation}
\vspace{-0.4\baselineskip}
\end{wrapfigure}
Figure~\ref{fig:ablation} reports a full-scale ablation on DSG across all four generators, three seeds, and 200 stratified prompts per seed, totaling 2,400 prompt--generator--seed evaluations for each variant. At a fixed image budget, full TARA reaches $82.8$; removing the explicit typed-expansion branch, the repair gate, the portfolio, or recovery lowers it by $3.8$, $3.6$, $2.2$, or $1.3$ points, respectively. The comparable first two drops show that gains arise jointly from explicit type-conditioned repair language and regression-controlled adoption, rather than from the gate alone.\looseness=-1

\emph{W/o explicit type branch} removes explicit type labels and the typed-expansion candidate while retaining the other portfolio branches---including minimal append---as well as fusion, recovery, the repair gate, and the matched image budget; \emph{w/o recovery} omits pruned relation/action recovery; and \emph{w/o portfolio} replaces typed repair compilation with a one-shot typed holistic rewrite. \emph{W/o repair gate} retains the complete repair pipeline and matched image budget but unconditionally adopts the single repaired image, isolating only the accept-or-revert policy.

\vspace{-0.55\baselineskip}
\section{Related Work}
\vspace{-0.3\baselineskip}

Diffusion models~\citep{ho2020denoising,song2020score,ddim02,diffusionbeatsgan03,vae08,rombach2022high,ho2022classifier,radford2021learning,t5-07}, scaled by stronger backbones such as SDXL~\citep{sdxl09}, SD~3~\citep{sd3-10}, and Flux~\citep{blackforestlabs2024flux}, and joined by autoregressive generators like Janus-Pro~\citep{chen2025januspro}, have made high-fidelity synthesis routine yet still fail on compositional and fine-grained semantic constraints, motivating prompt-side optimization without retraining or internal access.\looseness=-1

Early work targets visual appeal via style modifiers and keyword search~\citep{liu2023pretrain,sahoo2024systematic,liu2022design,pavlichenko2023best,taxonomy19}. Interactive systems~\citep{reprompt18,promptify23,promptmagician22,promptcharm20,promptexpansion21} and learned rewriters---Promptist~\citep{hao2023optimizing}, BeautifulPrompt~\citep{cao2023beautifulprompt}, NeuroPrompts~\citep{rosenman2024neuroprompts}, TIPO~\citep{yeh2024tipo}---automate prompt refinement but mainly improve aesthetics rather than semantic faithfulness. VisualPrompter~\citep{wu2026visualprompter} closes this gap by feeding VLM diagnostic feedback back into prompt rewriting, yet it applies one uniform expansion regardless of failure type. Under our formulation, such feedback-triggered rewriting is a single-operator allocation policy; TARA instead uses feedback to route each atomic failure to a type-conditioned intervention. The distinction is therefore not whether visual feedback is available, but how it is operationalized: prior methods decide \emph{what} to rewrite, whereas TARA additionally decides \emph{how each failure should be repaired}. Process-side methods~\citep{uffgtg24,manas2024opt2i,jiang2024frap,zhang2026freetext} improve alignment through multi-round generation or direct latent manipulation, requiring additional generator calls or internal access. TARA compiles its allocated local repairs into one prompt and gates one repaired output, achieving stronger accuracy within a single re-generation.\looseness=-1

Global metrics such as FID~\citep{fid40}, CLIPScore~\citep{hessel2021clipscore}, and learned preference scores~\citep{pickscore33,hps34} capture distributional or perceptual quality but cannot pinpoint which semantic requirements are met. Recent work decomposes prompts into atomic checkable questions~\citep{whatyousee27,stepbystepevaluation28,towardsqa29,factscore30}; DSG~\citep{cho2024dsg} and TIFA~\citep{hu2023tifa} instantiate this for T2I with dependency-aware question generation and VLM-based verification. TARA moves from feedback-triggered rewriting to feedback-routed repair: atomic feedback is not only an evaluation signal, but the routing variable that selects how each failed proposition is repaired.\looseness=-1

\vspace{-0.6\baselineskip}
\section{Conclusion}
\vspace{-0.3\baselineskip}
\emph{TARA} reformulates semantic prompt optimization as atomic repair allocation: diagnose, allocate, compile, and adopt. It leads all eight benchmark--generator cells ($+5.6$/$+2.6$ over VisualPrompter), preserves image quality, and runs fastest. More broadly, atomic visual feedback should not merely identify what failed; it should determine how each failure is repaired.\looseness=-1

\clearpage
\bibliography{iclr2026_conference}
\bibliographystyle{iclr2026_conference}

\clearpage
\appendix
\renewcommand{\contentsname}{Contents of the Paper}
{\hypersetup{linkcolor=black}\tableofcontents}
\bigskip

\section{Experimental Setup and Method Details}

This appendix provides full details: implementation and decoding settings, repair-language templates, the complete prompt templates used in portfolio rewriting and fusion (per-type rules, few-shot examples, fusion/minimal-append/uniform instructions), worked examples with step-by-step repair traces, qualitative comparisons across all generators and benchmarks, runtime accounting, threshold sensitivity, random-seed robustness, optimized-prompt-length and per-error-type repair-behavior analyses, and the complete VLM-as-Judge preference numbers. Unless otherwise specified, all semantic scores are percentages averaged over three seed groups with 200 prompts per seed group.

\subsection{Implementation and Runtime Protocol}\label{app:implementation}
All visual-feedback methods are run locally with the same OpenAI-compatible vLLM protocol for VLM diagnosis and LLM rewriting. The main endpoint serves Qwen3.5-9B under the alias \texttt{qwen3.5-9b}; distributed runs use local ports \texttt{8013}/\texttt{8023} for the same model, and the ablation/robustness runs use the same protocol on ports \texttt{8004}/\texttt{8026}. The endpoints are local-only (\texttt{127.0.0.1}) and are called with environment proxies disabled. The main semantic judge and prompt rewriter therefore share the same model family for VisualPrompter and TARA. The repair gate is executed per input before dataset-level aggregation, with one global rule fixed across all generators and benchmarks; no threshold or decision rule is tuned on the test set. After those outputs are fixed, Table~\ref{tab:judge_robust} independently re-scores them with Qwen3.5-4B, Qwen2.5-VL-7B-Instruct, Gemma4-31B, and Qwen3.5-27B evaluators that play no role in gate decisions.

For VLM yes/no diagnosis, the request contains the image and one benchmark atomic question only; it does not contain either the original prompt or the rewritten prompt. We use deterministic decoding for scoring (temperature $0$, maximum 8 output tokens for yes/no answers; auxiliary visual evidence, when enabled, is capped at 256 tokens). Prompt rewriting uses maximum 1024 new tokens. VisualPrompter follows its deterministic single-rewrite setting, while TARA samples text-only candidate rewrites with temperature $0.7$ and top-$p$ $0.8$ before fusing them into one final prompt. All random seeds are derived from the prompt id and seed group, and API sampling seeds are fixed per request for reproducibility. Table~\ref{tab:impl_settings} summarizes the image-generation and decoding settings used throughout.

\begin{table}[H]
\centering
\scriptsize
\caption{Image-generation and decoding settings used in the main experiments. SD models use 30 denoising steps and guidance 7.5 unless otherwise noted.}
\label{tab:impl_settings}
\begin{tabular}{l c c c c}
\toprule
Generator & Resolution & Steps & Guidance & Device dispatch \\
\midrule
Stable Diffusion v1.5 & $512\times512$ & 30 & 7.5 & CUDA:2 / CUDA:6 \\
Stable Diffusion v2.1 & $768\times768$ & 30 & 7.5 & CUDA:2 / CUDA:6 \\
Flux-dev & $1024\times1024$ & 28 & 3.5 & CUDA:2 / CUDA:6 \\
Janus-Pro & $384\times384$ & 576 & 5.0 & CUDA:2 / CUDA:6 \\
\bottomrule
\end{tabular}
\end{table}

Table~\ref{tab:runtime_accounting} accounts for the per-prompt cost of the two visual-feedback methods: both spend the same image budget, but TARA trades one long rewrite for several cheap text-only calls, ending with a shorter prompt and a lower repair time.

\begin{table}[H]
\centering
\scriptsize
\caption{Runtime accounting for the two visual-feedback methods. Image budget is matched; TARA spends more cheap text-only calls but keeps each rewrite short. ``Repair time'' is measured on prompts whose initial image triggers a rewrite.}
\label{tab:runtime_accounting}
\resizebox{\textwidth}{!}{%
\begin{tabular}{l c c c c c}
\toprule
Method & Image gens / prompt & Text LLM calls / prompt & Final words if repaired & No-repair time (s) & Repair time (s) \\
\midrule
VisualPrompter & 1.64 & 0.64 & 41.6 & 10.30 & 25.47 \\
TARA (Ours) & 1.64 & 3.37 & 22.7 & 10.15 & 19.37 \\
\bottomrule
\end{tabular}}
\end{table}

The end-to-end times in Table~\ref{tab:time} include image generation, VLM diagnosis, prompt rewriting, the final gate check, and disk I/O for saved records. They exclude only one-time model loading. Because all methods use the same generated images for scoring and the same local evaluation endpoint, the runtime comparison measures the optimizer pipeline rather than differences in evaluator deployment.

\subsection{Repair Mapping and Templates}\label{app:templates}
TARA maps atomic-evaluation categories to eight repairable error types, each associated with a distinct local language transformation. The mapping is fixed before evaluation and shared across all generators and benchmarks. Table~\ref{tab:mapping_templates} lists the exact categories and representative templates.

\begin{table}[H]
\centering
\scriptsize
\caption{Category-to-type mapping and repair-language templates used by TARA. Bracketed fields are filled from the failed atomic proposition; already-correct entities and attributes are explicitly preserved in the LLM instruction.}
\label{tab:mapping_templates}
\resizebox{\textwidth}{!}{%
\begin{tabular}{l l p{0.58\textwidth}}
\toprule
DSG category & Repair type & Repair language used in candidate prompts \\
\midrule
Entity & Missing object & Make \texttt{[object]} clearly visible and salient in the scene; keep its requested attributes and relation anchors. \\
Count & Wrong count & Show exactly \texttt{[number]} distinct \texttt{[objects]}, separated and countable; avoid extra or missing instances. \\
Attribute & Wrong attribute & Bind \texttt{[attribute]} directly to \texttt{[object]}; avoid transferring the attribute to other objects. \\
Relation & Wrong relation & Place \texttt{[subject]} unambiguously \texttt{[relation]} \texttt{[object]} with clear spatial anchors such as left/right/above/below/inside. \\
Action & Wrong action & Depict \texttt{[subject]} visibly performing \texttt{[action]} with pose, contact, or motion cues involving \texttt{[object]} when present. \\
Text & Missing or illegible text & Render the exact quoted string \texttt{[text]} in large, legible letters on the requested carrier (sign, label, smoke trail, etc.). \\
Style & Style mismatch & Strengthen the required style (medium, lighting, mood, era) without changing any scene content, objects, counts, or attributes. \\
Global & Global mismatch & Preserve the concrete scene while enforcing the requested global condition, viewpoint, or scene-level constraint. \\
\bottomrule
\end{tabular}}
\end{table}

These templates only shape the rewritten prompt. During evaluation, the VLM judge receives neither the template nor the optimized prompt, only the image and the original benchmark question. This separation is important: type-aware language can help the image generator attend to the missing semantics, but it cannot directly hint the automatic yes/no evaluator.

\subsection{Full Algorithm}
Algorithm~\ref{alg:tara} states the complete single-pass procedure illustrated in Figure~\ref{fig:pipeline}.

\begin{algorithm}[H]
\caption{TARA: single-pass atomic repair allocation}
\label{alg:tara}
\begin{algorithmic}[1]
\Require prompt $p$, generator $G$, judge $V$, propositions $Q(p)$, threshold $\tau{=}0.72$
\State $x_0 \gets G(p)$;\enspace $d_0 \gets \textsc{Diagnose}(x_0, Q(p), V)$
\If{$d_0$ is perfect}
    \State \Return $(p, x_0)$ \Comment{already correct: one generation}
\EndIf
\State $F \gets \textsc{RepairTargets}(d_0)$ \Comment{failures $+$ pruned wrong relations/actions}
\State $C \gets \{\textsc{Uniform}, \textsc{Typed}, \textsc{MinimalAppend}\}(p, F)$ \Comment{text-only}
\If{$s(x_0) \ge \tau$}
    \State $C \gets C \cup \{\textsc{ReSeed}(p)\}$
\EndIf
\State $\hat{p} \gets \textsc{LLMFuse}(C, F, p)$ \Comment{text-only: one fused prompt}
\State $\hat{x} \gets G(\hat{p})$;\enspace $\hat{d} \gets \textsc{Diagnose}(\hat{x}, Q(p), V)$ \Comment{$\hat{d}$ yields $s(\hat{x})$}
\If{$s(\hat{x}) > s(x_0)$}
    \State \Return $(\hat{p}, \hat{x})$ \Comment{accept: two generations}
\Else
    \State \Return $(p, x_0)$ \Comment{reject and revert}
\EndIf
\end{algorithmic}
\end{algorithm}

\subsection{Prompt Templates}\label{app:prompts}

Below we list the exact prompt templates used by TARA's portfolio rewriting and fusion stages.
Bracketed fields (e.g.\ \texttt{\{original\_prompt\}}) are filled at runtime from the diagnosed prompt and its failed atomic propositions.

\paragraph{Per-type repair rules.}
Each error type maps to a type-specific system instruction that encodes domain knowledge about how diffusion models fail on that type (Table~\ref{tab:type_rules}).

\begin{table}[H]
\centering
\scriptsize
\caption{Per-type repair rules used in the typed-expansion candidate. Each rule is the system instruction given to the LLM rewriter for that failure type.}
\label{tab:type_rules}
\renewcommand{\arraystretch}{1.3}
\begin{tabular}{l p{11.5cm}}
\toprule
Error type & Repair rule (system instruction to LLM) \\
\midrule
Missing object &
Rewrite the prompt so each missing object becomes an explicit, concrete subject of the scene and is made visually salient: name it directly, give it a short concrete description, and place it in the foreground rather than merely implying it. \\
Wrong count &
Pin the exact quantity hard: state the number as BOTH a word and a digit, describe the instances as separate and individually visible (e.g.\ ``exactly three cats (3), three separate cats, evenly spaced, fully visible, not overlapping''), and avoid vague plurals. \\
Wrong attribute &
Bind each attribute tightly to ITS object: put the adjective immediately before the noun, repeat the binding once for emphasis, and leave every other object's attributes unchanged (e.g.\ ``a bright red car, the car painted solid red''). \\
Wrong relation &
Make the relation explicit with concrete positional/action language, and describe a simple layout that anchors BOTH objects (e.g.\ ``a lamp on the left and a sofa on the right, the lamp clearly to the left of the sofa''). \\
Wrong action &
Depict the subject visibly performing the action: make pose, contact, and motion cues explicit and keep the interacting object present; do not replace the action with a static description. \\
Style mismatch &
Strengthen the required style with concrete descriptors (medium, lighting, mood, era/artist if implied) WITHOUT changing any scene content, objects, counts, or attributes. \\
Missing text &
Quote the exact string and request it to be legible (e.g.\ ``the word `OPEN' written clearly in large bold letters''). Keep the text short and exact; do not paraphrase it. \\
Global mismatch &
Enforce the requested viewpoint, scene-level condition, or overall composition explicitly, while keeping all local objects, counts, and attributes unchanged. \\
\bottomrule
\end{tabular}
\end{table}

\paragraph{Global constraints.}
Every rewrite candidate receives the following constraint block appended to its type-specific rule:

\begin{quote}
\small
\texttt{Rules: fix ONLY the issues listed below; keep everything already correct unchanged; do NOT add objects that are not in the original prompt or the issues; preserve the user's intent and overall scene; output exactly ONE image prompt on a single line --- no quotes, no list, no explanation.}
\end{quote}

\paragraph{Few-shot examples.}
Representative type-specific instructions include one or two few-shot examples; Table~\ref{tab:fewshot} lists the examples used.

\begin{table}[H]
\centering
\scriptsize
\caption{Representative few-shot examples used in the type-specific rewrite instructions.}
\label{tab:fewshot}
\renewcommand{\arraystretch}{1.25}
\begin{tabular}{l p{4cm} p{7.5cm}}
\toprule
Type & Issue & Rewritten prompt \\
\midrule
Missing object & the oven is missing & a cozy kitchen with a large stainless-steel oven clearly visible in the foreground \\
Wrong count & there must be exactly three cats & exactly three cats (3) sitting on a sofa, three separate cats evenly spaced and fully visible, not overlapping \\
Wrong count & there must be five balloons & exactly five balloons (5) floating in the sky, five distinct balloons clearly separated \\
Wrong attribute & the car should be red & a man standing next to a bright red car, the car painted solid red \\
Wrong relation & the cat should be inside the box & a cat sitting inside an open cardboard box, the cat fully contained within the box \\
Wrong relation & the lamp should be to the left of the sofa & a room with a lamp on the left side and a sofa on the right side, the lamp clearly to the left of the sofa \\
Style mismatch & it should look cinematic & a street at night, cinematic film still, dramatic lighting, moody atmosphere, deep shadows \\
Missing text & the sign should read OPEN & a shop sign on a wall with the word ``OPEN'' written clearly and legibly in large bold letters \\
\bottomrule
\end{tabular}
\end{table}

\paragraph{Fusion instruction.}
After the portfolio candidates are generated, TARA fuses them with the following LLM instruction:

\begin{quote}
\small
\texttt{You are selecting the final prompt for a text-to-image generator. There is only ONE remaining image generation, so choose the prompt most likely to satisfy the semantic questions.}

\texttt{Original prompt: \{original\_prompt\}}

\texttt{Already-correct aspects to preserve: \{passed\}}

\texttt{Failed aspects to fix: \{failed\}}

\texttt{Candidate prompts: \{choices\}}

\texttt{Decision rule: Prefer semantic correctness over aesthetics. Penalize prompts that drop original entities, add unrelated objects, make counts ambiguous, or describe relations/actions vaguely.}

\texttt{Task: Synthesize one final image prompt using the strongest parts of the candidates. Keep the original intent, preserve already-correct aspects, fix all failed aspects, avoid unrelated objects and generic quality filler, and stay under 70 words if possible. Output exactly one final image prompt, no explanation.}
\end{quote}

\paragraph{Minimal-append instruction.}
The conservative candidate keeps the original prompt nearly verbatim and appends only the shortest typed clauses:

\begin{quote}
\small
\texttt{Keep the original prompt almost verbatim. Append only the shortest concrete phrases needed to fix the failed aspects. Do not rewrite correct content, do not add unrelated objects, and do not add generic aesthetic filler.}

\texttt{Type rules: For count, use exact word + digit and separate visible instances. For relation, add a clear left/right/above/below layout. For action, add visible pose/contact/motion. For text, quote the exact legible text. For object/attribute/global/style, add only the missing concrete phrase.}
\end{quote}

\paragraph{Uniform-expansion instruction (in-method control).}
The type-agnostic uniform-expansion candidate receives no type information:

\begin{quote}
\small
\texttt{You repair text-to-image prompts. The aspects below were requested but are missing or incorrect in the generated image. Rewrite the prompt by enriching it with concrete detail so that ALL listed aspects are clearly depicted at once, while preserving the original intent and everything already correct. Output exactly ONE image prompt on a single line, no explanation.}
\end{quote}

\subsection{Worked Example}
\label{app:worked}
\begin{figure}[htbp]
\centering
\begin{minipage}[b]{0.26\textwidth}\centering
\includegraphics[width=\linewidth]{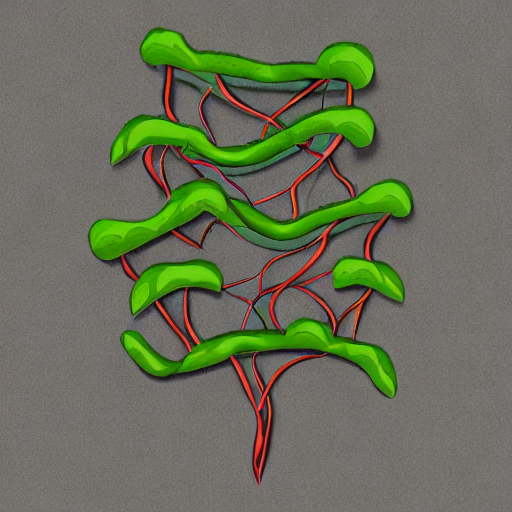}\\[2pt]
{\scriptsize Initial render, score $0.00$}
\end{minipage}\hspace{0.015\textwidth}
\begin{minipage}[b]{0.26\textwidth}\centering
\setlength{\fboxsep}{0pt}\setlength{\fboxrule}{2pt}%
\fcolorbox{cPos}{white}{\includegraphics[width=\linewidth]{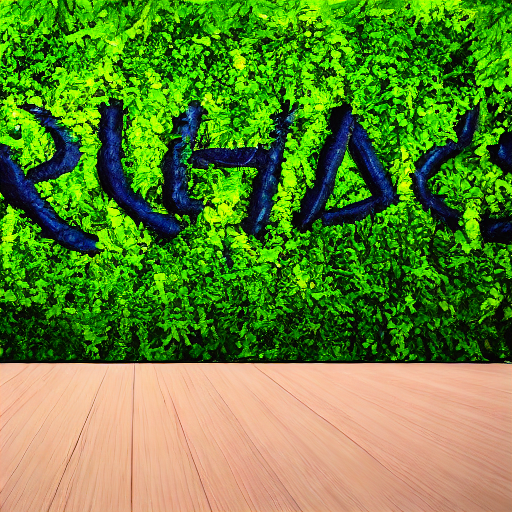}}\\[2pt]
{\scriptsize TARA render, score $\mathbf{0.83}$}
\end{minipage}\hspace{0.015\textwidth}
\begin{minipage}[b]{0.40\textwidth}\scriptsize
\textbf{Original:} studio shot of vines in the shape of text knowledge is power sprouting, centered\\[4pt]
\textbf{TARA (one fused rewrite, one regeneration):} a studio shot of lush green vines clearly shaped into the large, legible text ``knowledge is power'' sprouting from the ground and centered in the frame
\end{minipage}

\vspace{7pt}
\scriptsize
\begin{tabular}{@{}c l l l l@{}}
\toprule
\# & Atomic proposition (DSG question) & Category & Initial & Routed repair operator \\
\midrule
1 & Is this a studio shot? & global & \textcolor{cNeg}{fail} & \texttt{ENFORCE\_GLOBAL} \\
2 & Are there vines? & entity & \textcolor{cNeg}{fail} & \texttt{ADD\_OBJECT} \\
3 & Are the vines in the shape of text? & attribute & \textcolor{gray}{pass} & --- \\
4 & Does the text say ``knowledge is power''? & text & \textcolor{cNeg}{fail} & \texttt{ENFORCE\_TEXT} \\
5 & Are the vines sprouting? & attribute & \textcolor{gray}{pass} & --- \\
6 & Are the vines centered? & attribute & \textcolor{gray}{pass} & --- \\
\bottomrule
\end{tabular}
\caption{Worked example of type-aware repair (SD~1.5, DrawText). The initial image is decomposed into DSG atomic propositions and judged by the VLM; the three failed propositions are each routed to type-specific operators (global condition, missing object, and rendered text). TARA fuses the three typed repairs into a \emph{single} rewritten prompt, regenerates once, and the semantic score rises from $0.00$ to $0.83$---all within the single-pass image budget.}
\label{fig:worked}
\end{figure}

Figure~\ref{fig:worked} traces one prompt through TARA's pipeline: the initial render, the per-proposition diagnosis with assigned error types, the typed repairs fused into a single rewrite, and the regenerated image---using the category-to-type templates of Table~\ref{tab:mapping_templates}.

\paragraph{Additional worked examples (text-only).}
Below we show two further cases where TARA's typed repair produces a large score jump, one for \emph{wrong count} and one for \emph{wrong relation}. For post-hoc analysis only, we additionally rendered each candidate's output in isolation; these renders are not part of TARA's inference and are excluded from the main image budget. In each case no individually rendered candidate improves the score, while the fused prompt combines the strongest typed clauses and jumps sharply.

\textbf{Count repair (SD~2.1, DSG, \texttt{localized\_narratives\_59}).}
Original: ``In this image I can see few white color flowers. In the background I can see few leaves.''
Initial score: $0.67$. The separately rendered uniform, typed, minimal-append, and heuristic-selector outputs all score $0.67$; the fused prompt---``In this image I can see exactly two white flowers and exactly five leaves in the background, all clearly visible and distinct''---scores $\mathbf{1.00}$, fixing both vague counts.

\textbf{Relation repair (SD~2.1, DSG, \texttt{vrd\_42}).}
Original: ``person sit on chair. chair on the right of monitor. vase on table. pot on table.''
Initial score: $0.55$. Again, the separately rendered outputs stay at $0.55$; fusion yields ``A person sits on a chair positioned to the right of a monitor, with a vase and a pot placed on a table in the scene''---score $\mathbf{0.91}$, resolving ambiguous spatial relations.

\section{Supplementary Qualitative Results}

\subsection{Qualitative Comparisons}
\begin{figure}[htbp]
\centering
\includegraphics[width=\textwidth]{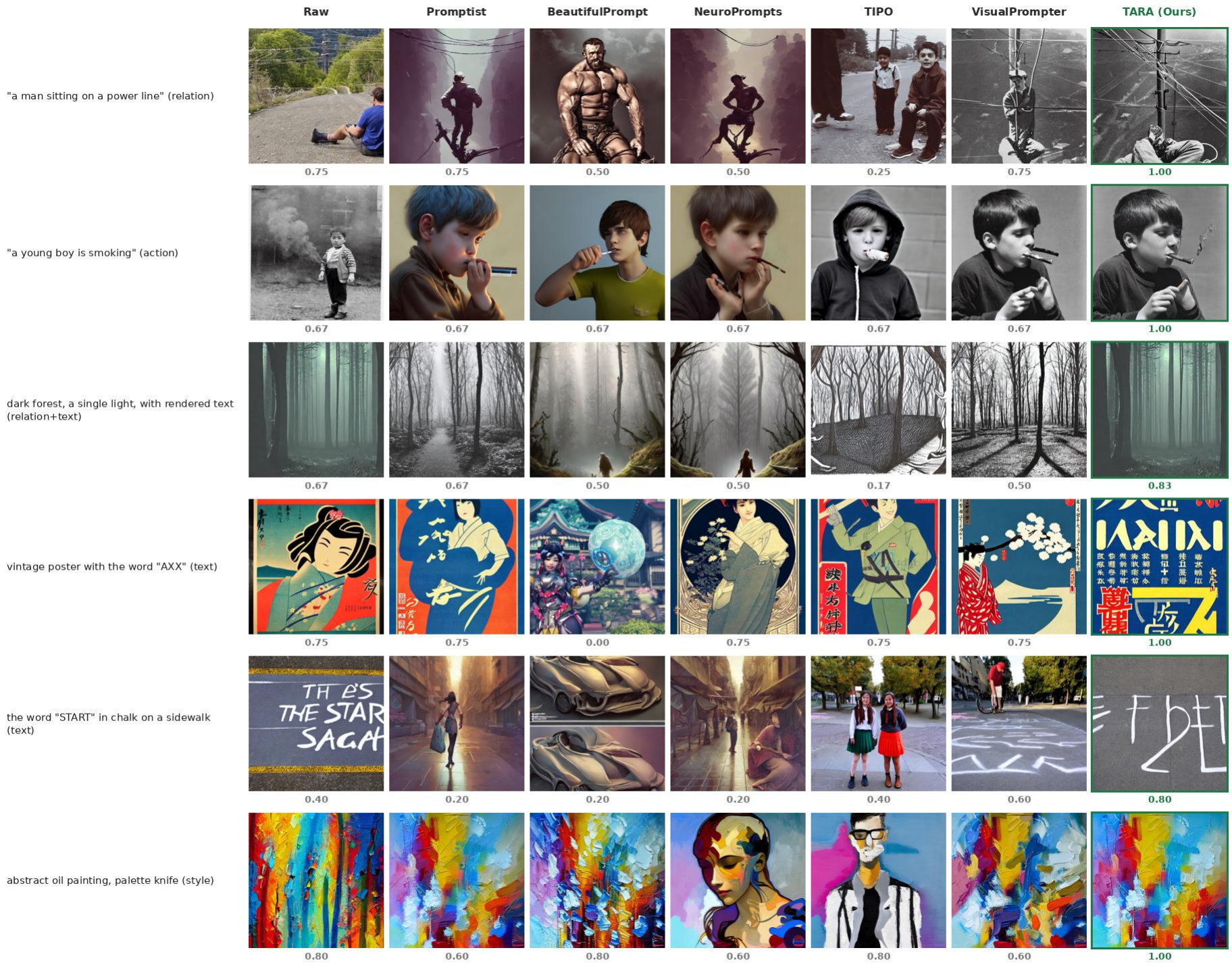}
\caption{Qualitative comparison on SD~1.5 / DSG. Each row is one prompt; columns are the raw prompt, four prompt-optimization baselines, the official VisualPrompter, and TARA (green outline). The number below each image is its VLM semantic score. TARA satisfies all atomic requirements---relations, actions, rendered text, and global style---where the baselines leave at least one unmet.}
\label{fig:qual_appendix}
\end{figure}

Figure~\ref{fig:qual_appendix} shows representative cases on Stable Diffusion~v1.5 (DSG) where every prompt-optimization baseline---including the official VisualPrompter---fails to satisfy the prompt while TARA renders an image that does. The number below each image is its VLM semantic score for that prompt.

\subsection{Per-Generator Galleries}
\label{app:galleries}
\begin{figure}[p]\centering
\includegraphics[width=\textwidth]{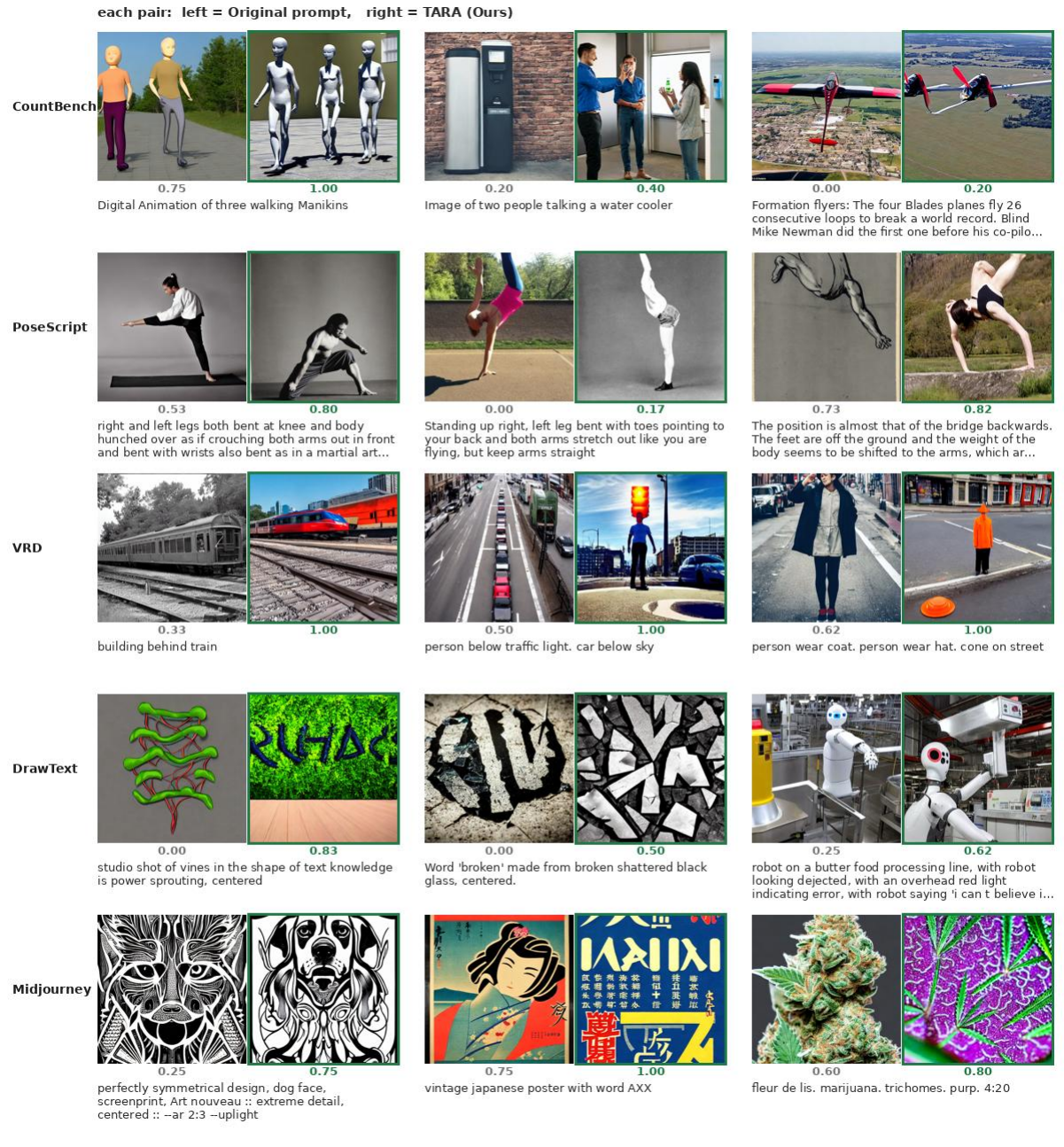}
\caption{Original vs.\ TARA on Stable Diffusion~v1.5 (DSG), grouped by source category. Left: original prompt; right (green outline): TARA. Numbers are VLM semantic scores.}
\label{fig:gal_dsg_sd15}
\end{figure}

\begin{figure}[p]\centering
\includegraphics[width=\textwidth]{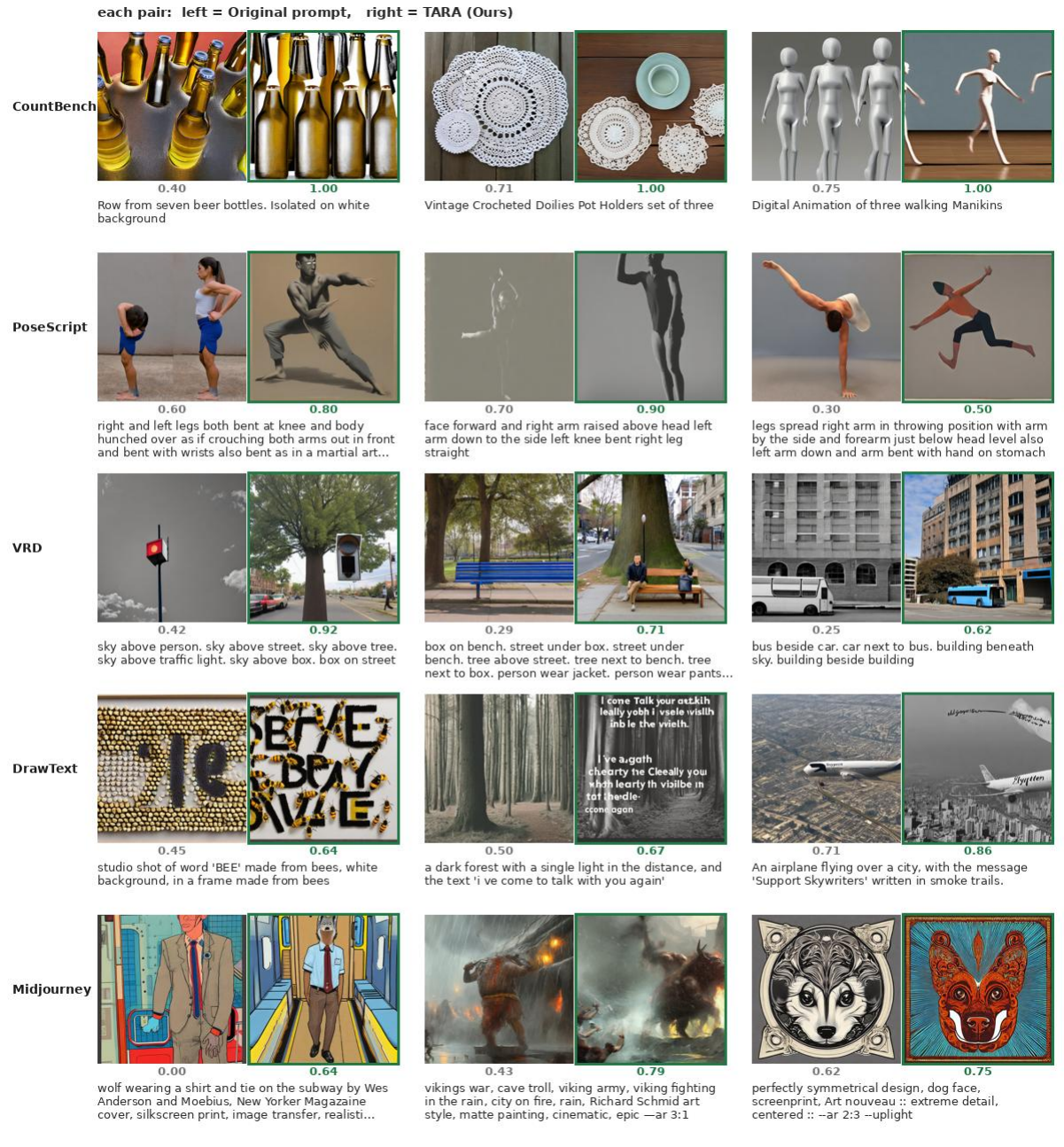}
\caption{Original vs.\ TARA on Stable Diffusion~v2.1 (DSG). Conventions as in Figure~\ref{fig:gal_dsg_sd15}.}
\label{fig:gal_dsg_sd21}
\end{figure}

\begin{figure}[p]\centering
\includegraphics[width=\textwidth]{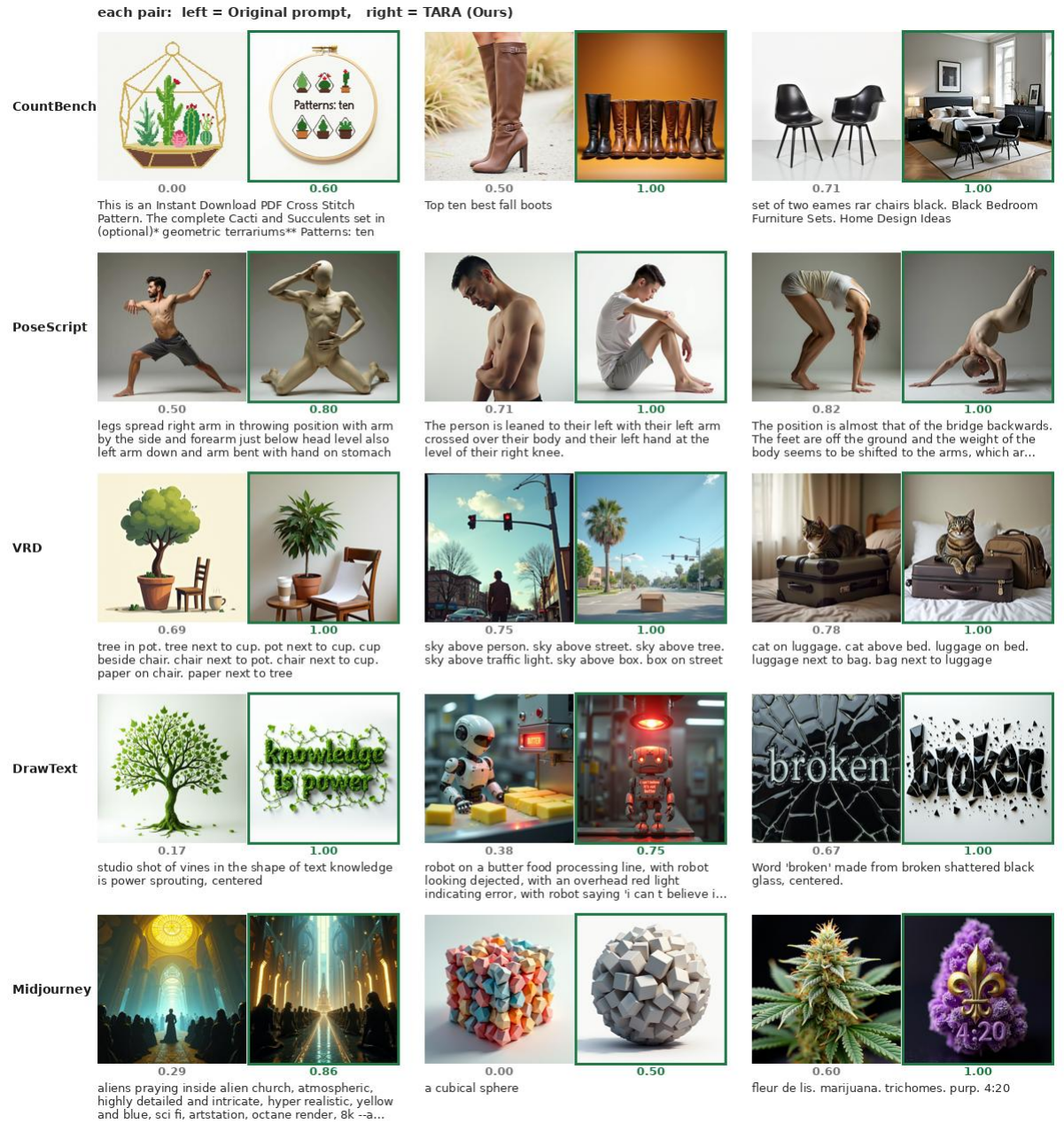}
\caption{Original vs.\ TARA on FLUX-dev (DSG). Conventions as in Figure~\ref{fig:gal_dsg_sd15}.}
\label{fig:gal_dsg_flux}
\end{figure}

\begin{figure}[p]\centering
\includegraphics[width=\textwidth]{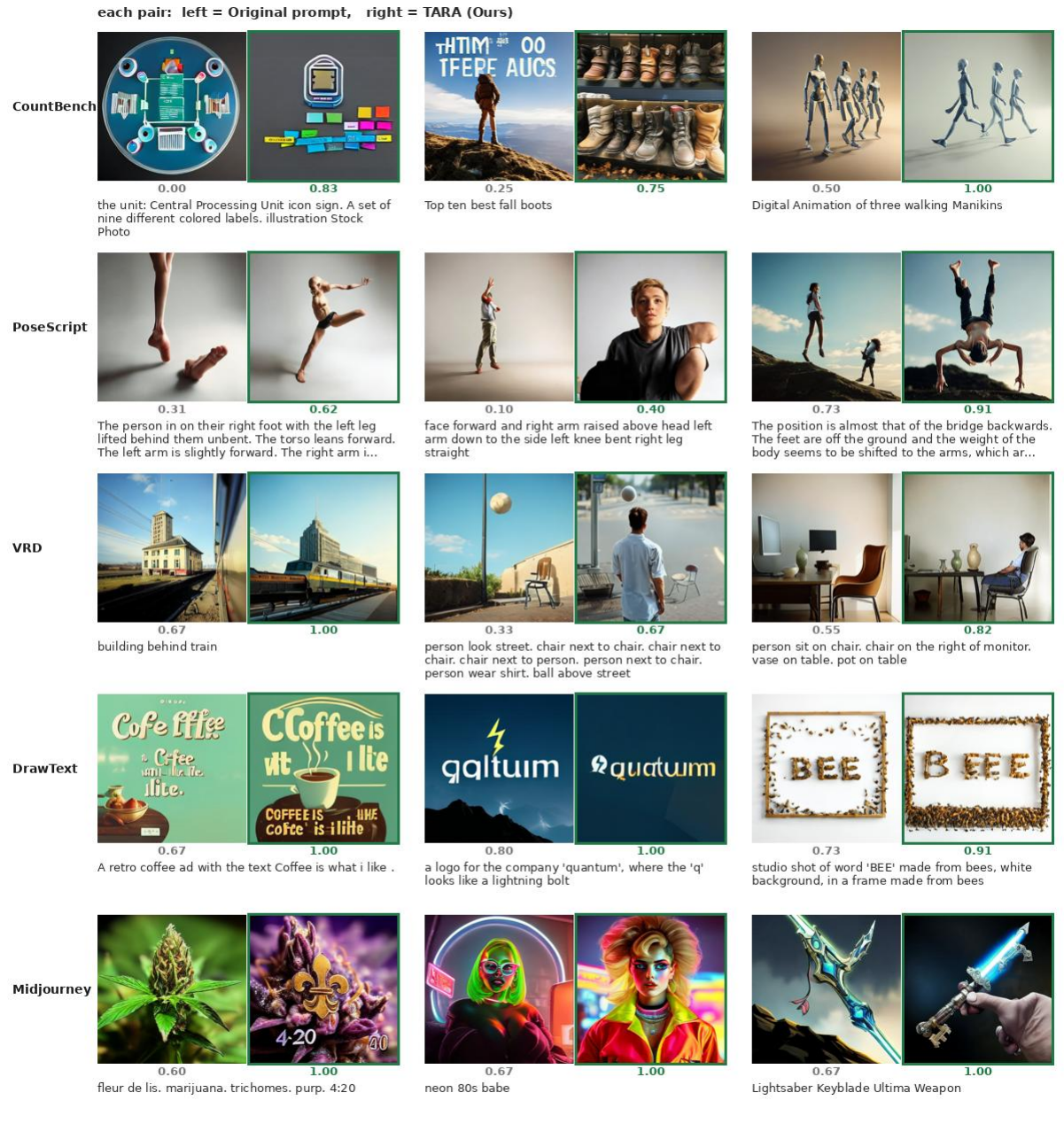}
\caption{Original vs.\ TARA on Janus-Pro (DSG). Conventions as in Figure~\ref{fig:gal_dsg_sd15}.}
\label{fig:gal_dsg_janus}
\end{figure}

\begin{figure}[p]\centering
\includegraphics[width=\textwidth]{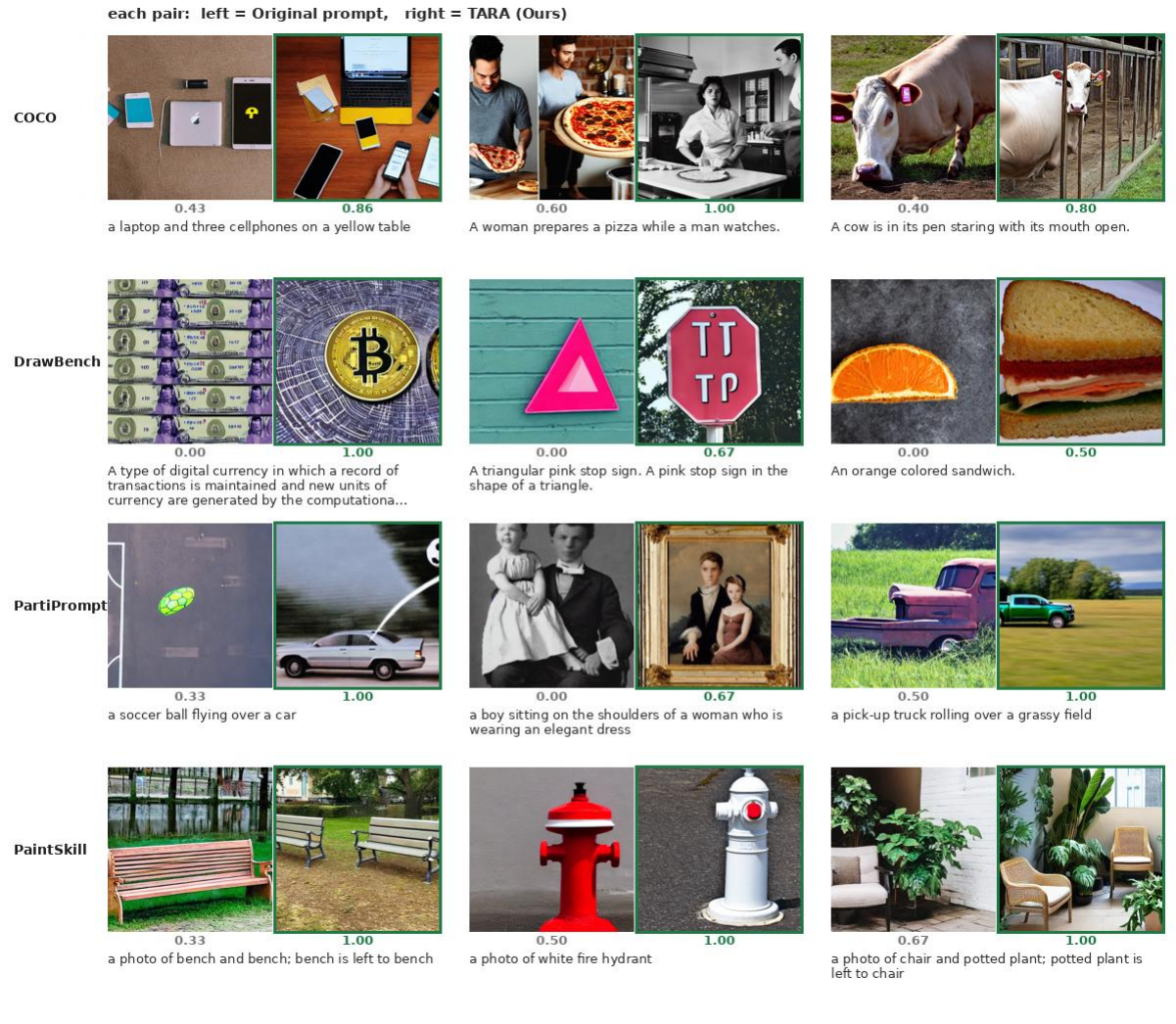}
\caption{Original vs.\ TARA on Stable Diffusion~v1.5 (TIFA), grouped by source category. Conventions as in Figure~\ref{fig:gal_dsg_sd15}.}
\label{fig:gal_tifa_sd15}
\end{figure}

\begin{figure}[p]\centering
\includegraphics[width=\textwidth]{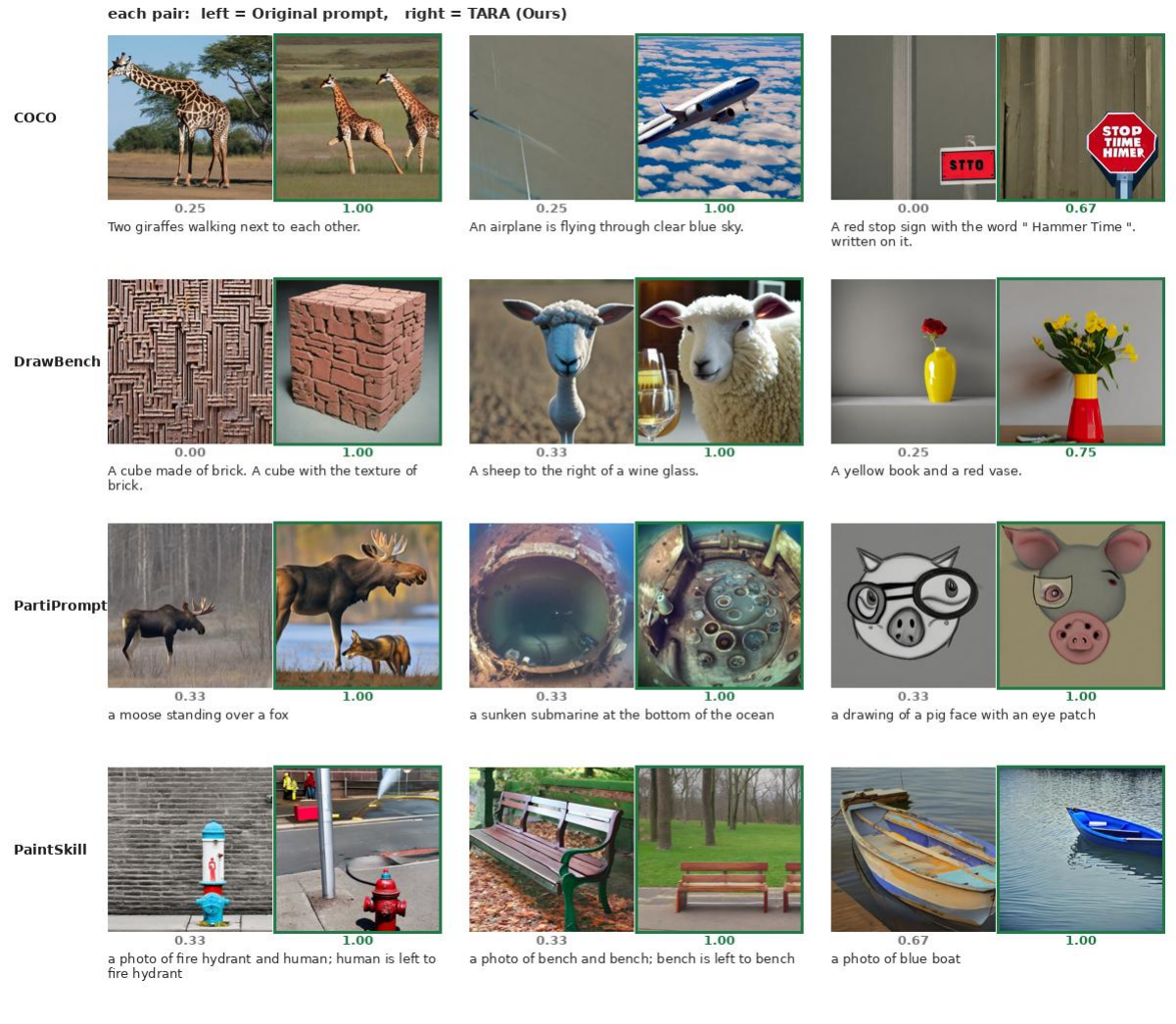}
\caption{Original vs.\ TARA on Stable Diffusion~v2.1 (TIFA). Conventions as in Figure~\ref{fig:gal_dsg_sd15}.}
\label{fig:gal_tifa_sd21}
\end{figure}

\begin{figure}[p]\centering
\includegraphics[width=\textwidth]{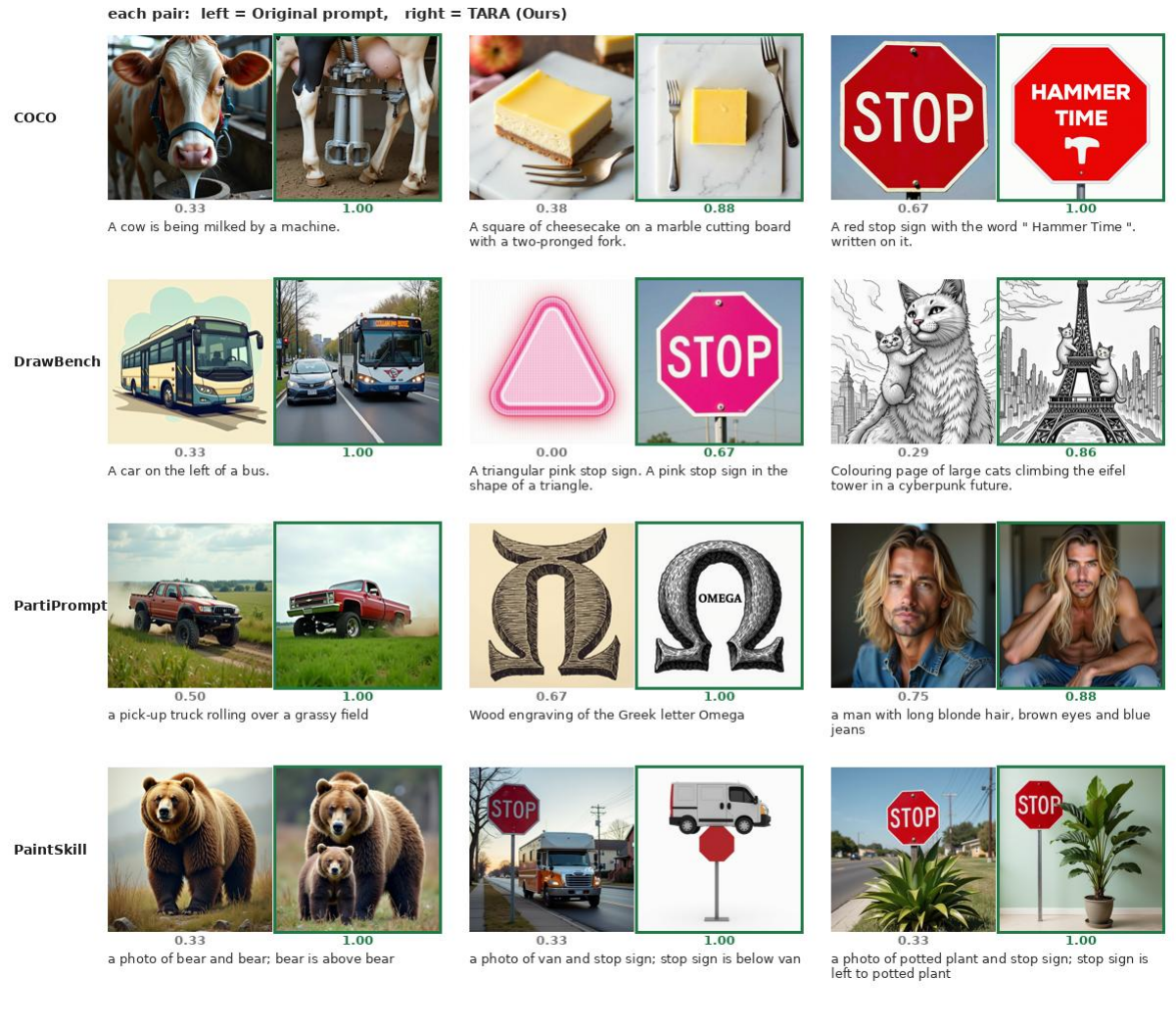}
\caption{Original vs.\ TARA on FLUX-dev (TIFA). Conventions as in Figure~\ref{fig:gal_dsg_sd15}.}
\label{fig:gal_tifa_flux}
\end{figure}

\begin{figure}[p]\centering
\includegraphics[width=\textwidth]{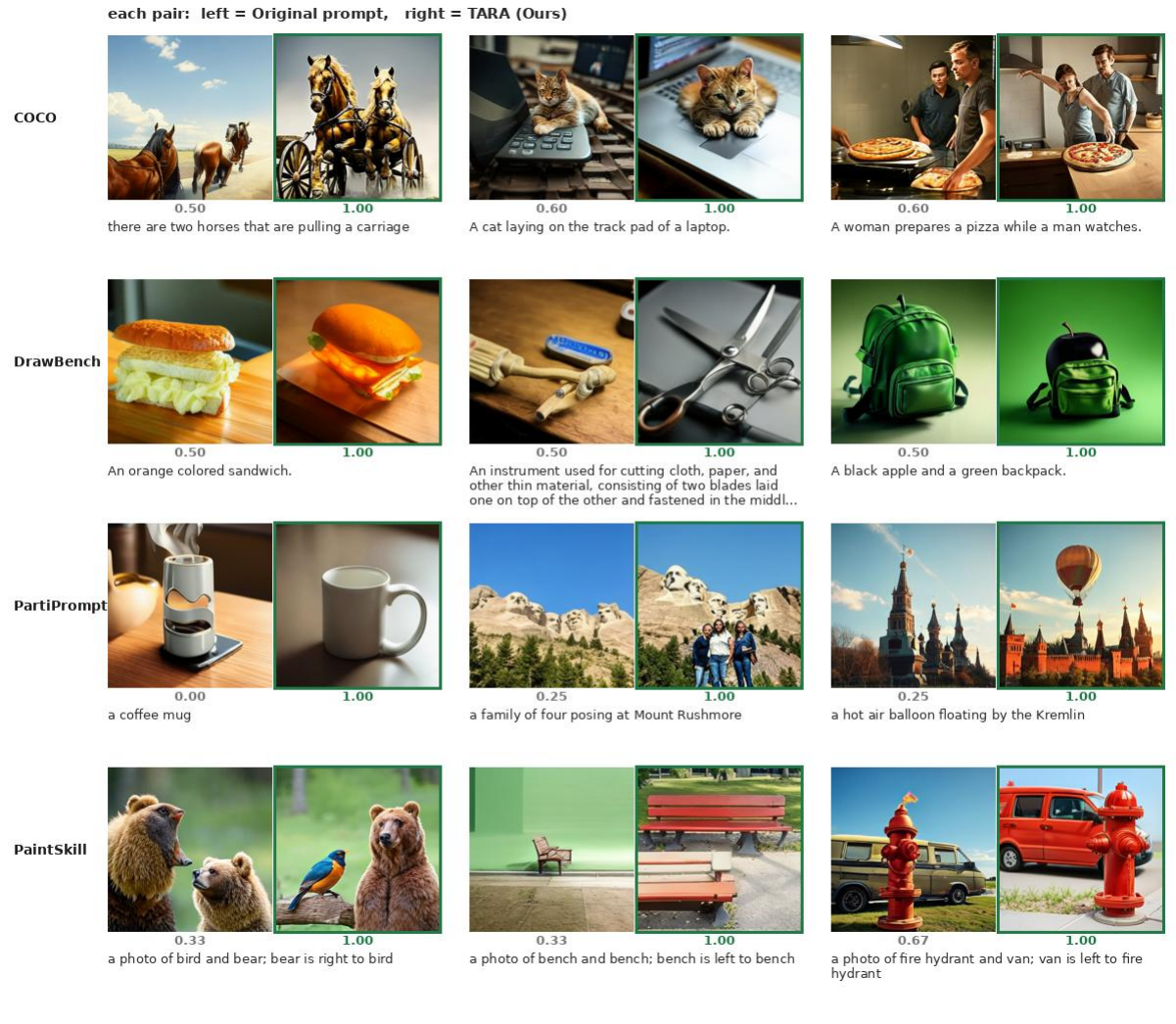}
\caption{Original vs.\ TARA on Janus-Pro (TIFA). Conventions as in Figure~\ref{fig:gal_dsg_sd15}.}
\label{fig:gal_tifa_janus}
\end{figure}

Figures~\ref{fig:gal_dsg_sd15}--\ref{fig:gal_tifa_janus} show, for each generator and both benchmarks (DSG and TIFA), original vs.\ TARA-optimized images grouped by source category. In every pair the left image is rendered from the original prompt and the right (green outline) from TARA's optimized prompt; the numbers are VLM semantic scores. The gains are largest exactly where typed repair targets---counts, relations and poses, and rendered text---and they transfer across both benchmarks.

\clearpage

\subsection{Adaptation to Commercial Online Generators}
\label{app:online}
\begin{figure}[htbp]
\centering
\setlength{\tabcolsep}{4pt}
\renewcommand{\arraystretch}{1.0}
\begin{tabular}{@{}*{3}{>{\centering\arraybackslash}p{0.315\textwidth}}@{}}
{\small\bfseries Original} & {\small\bfseries VisualPrompter} & {\small\bfseries TARA (Ours)} \\
\includegraphics[width=\linewidth]{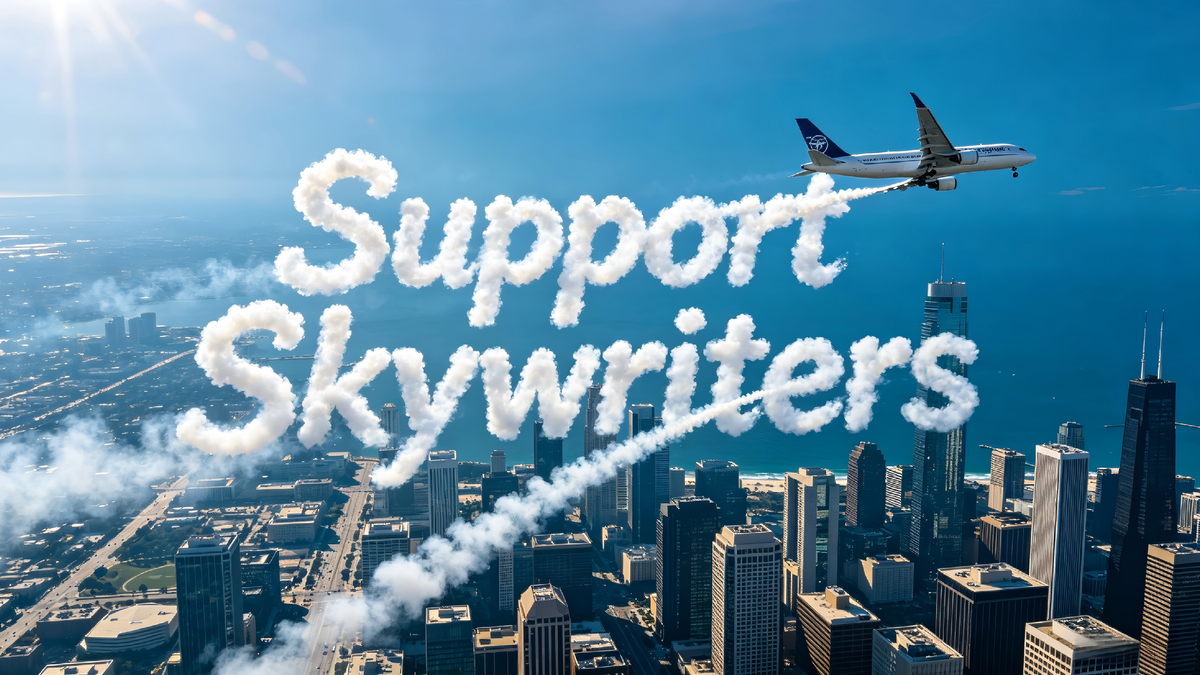} &
\includegraphics[width=\linewidth]{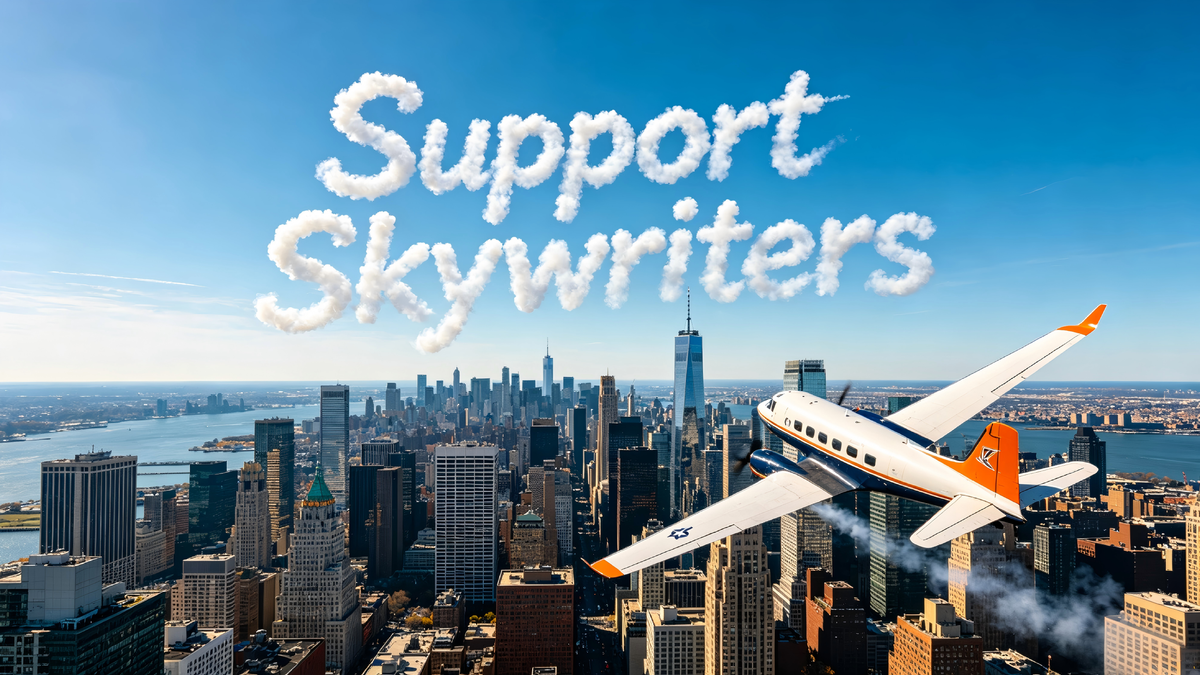} &
\includegraphics[width=\linewidth]{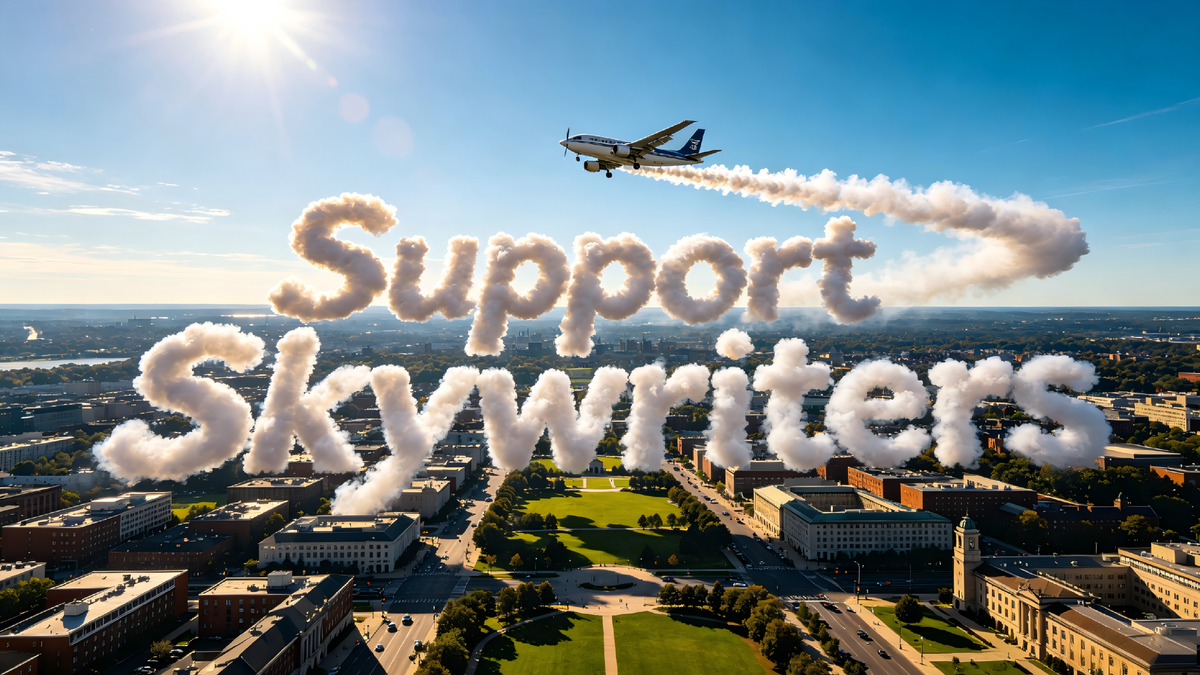} \\
{\footnotesize An airplane flying over a city, with the message `Support Skywriters' written in smoke trails.} &
{\footnotesize An airplane is flying over a city, with white smoke above the city. The smoke creates the message ``Support Skywriters'' in the sky. Tall buildings are in the city.} &
{\footnotesize An airplane flying over a city, with the message `Support Skywriters' clearly written in large, legible smoke trails.} \\[4pt]
\includegraphics[width=\linewidth]{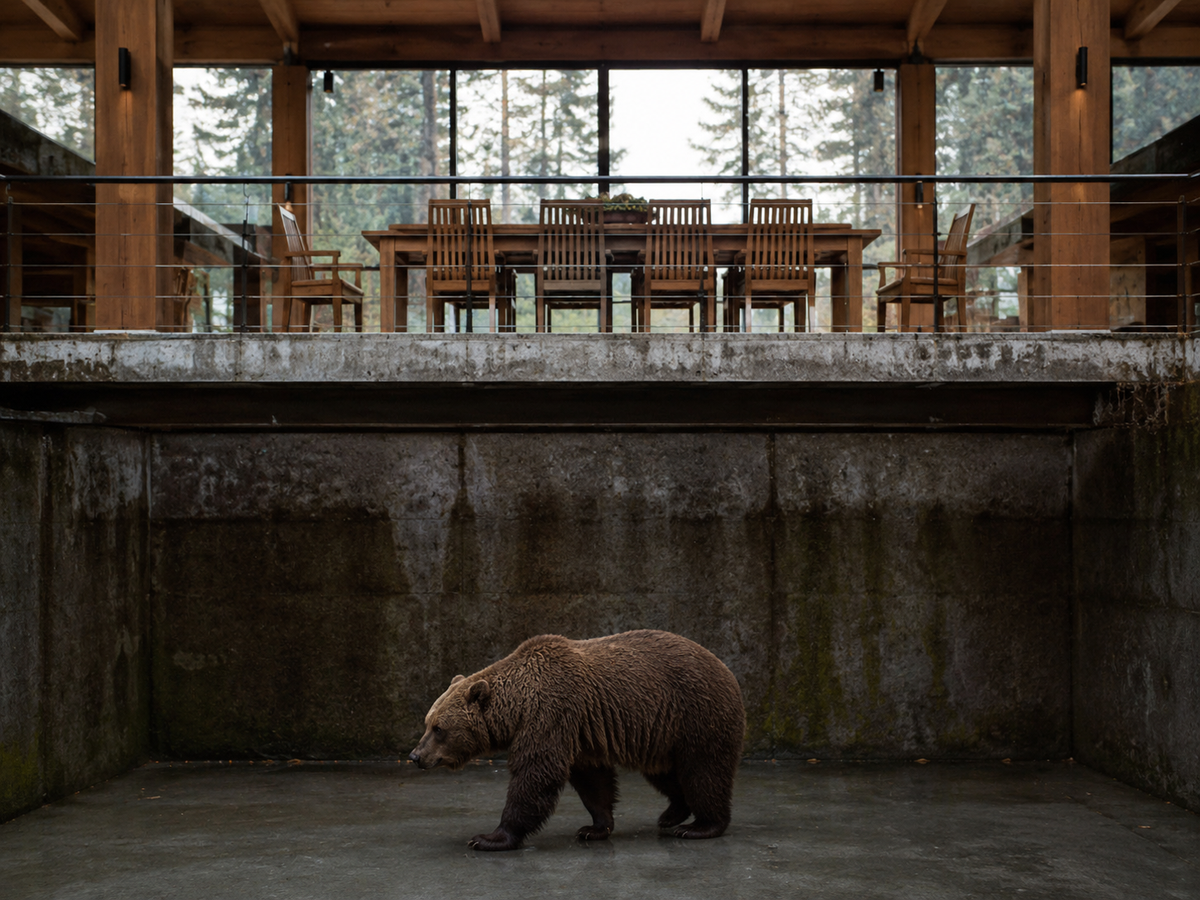} &
\includegraphics[width=\linewidth]{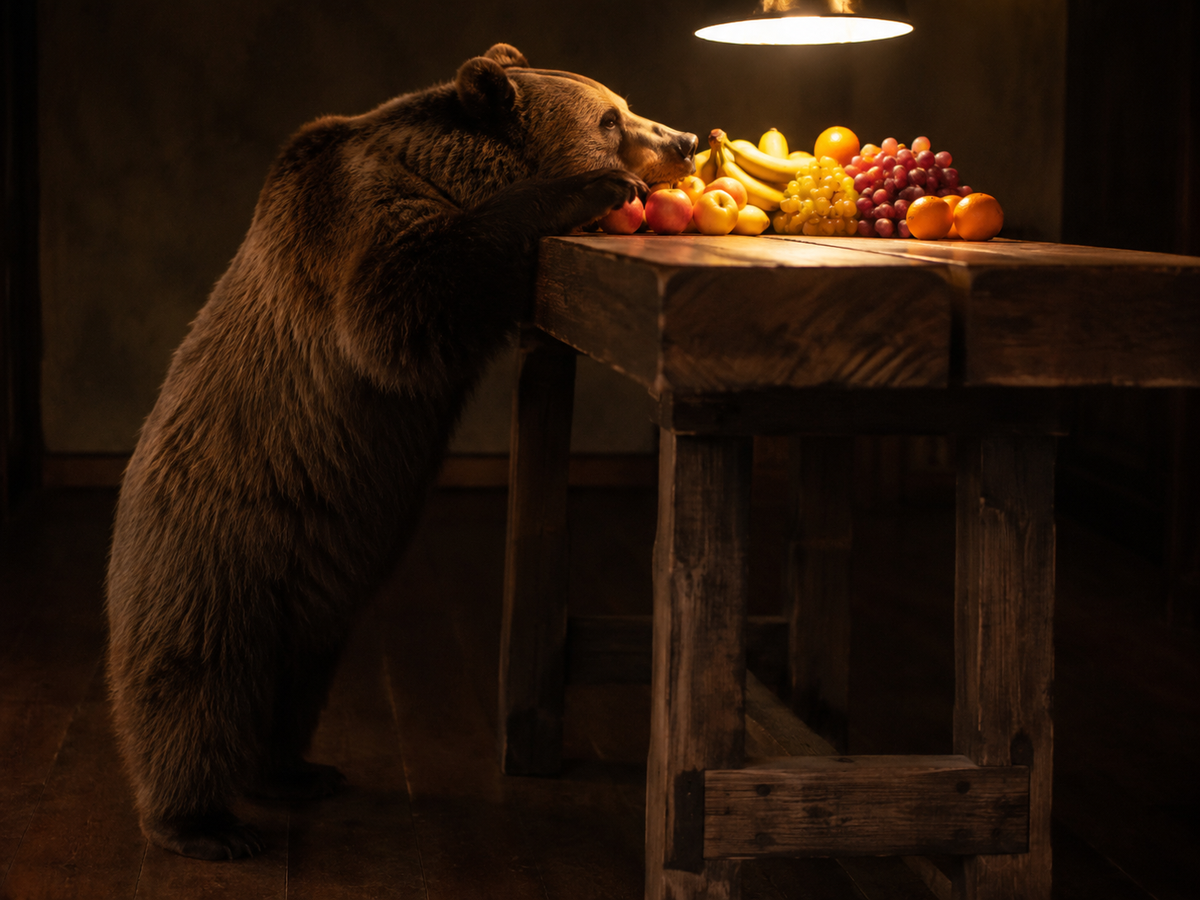} &
\includegraphics[width=\linewidth]{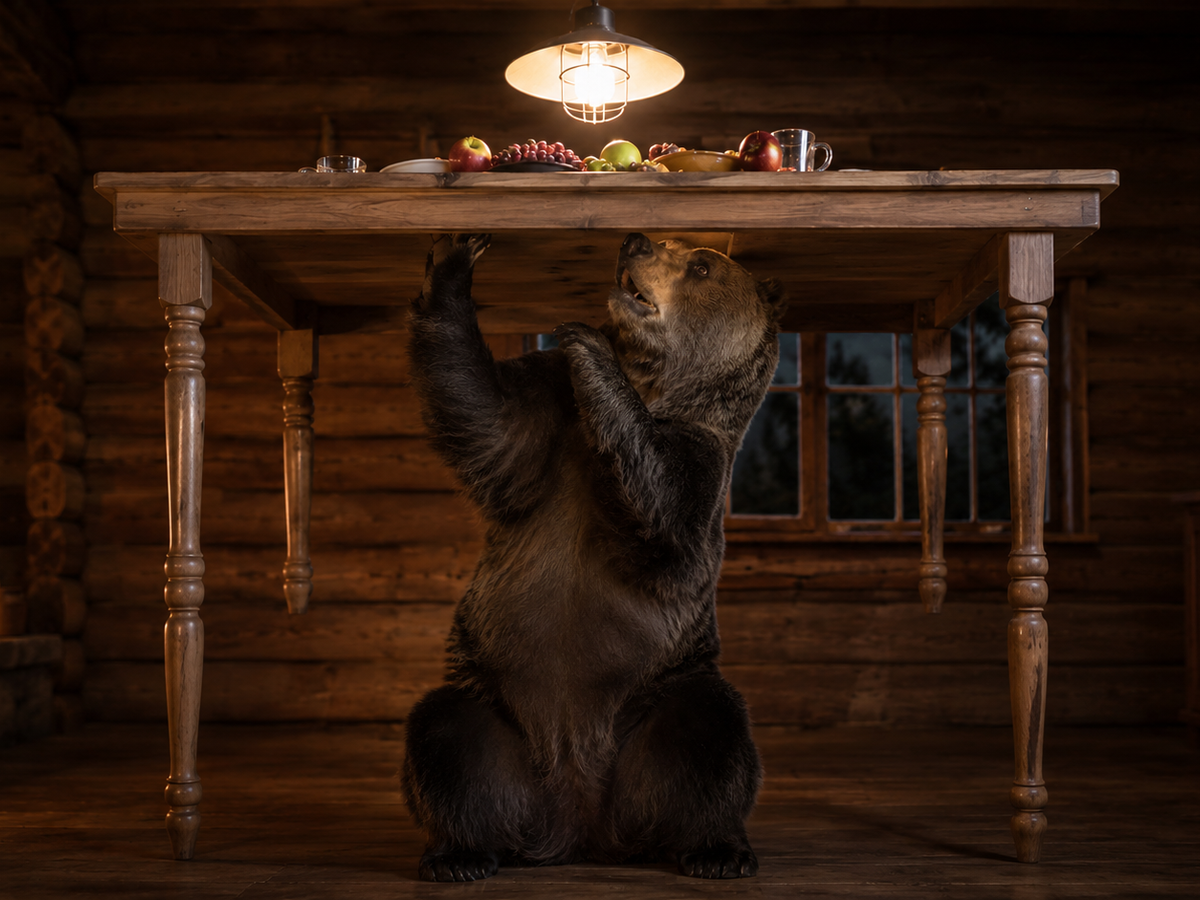} \\
{\footnotesize a photo of bear and dining table; dining table is above bear} &
{\footnotesize a photo of wooden dining table and bear; dining table is above bear with its light on. There are some fruits on the dining table, and the bear is eating them on the table.} &
{\footnotesize a photo of a bear and a dining table where the dining table is positioned above the bear, with the bear located below the table.} \\
\end{tabular}
\caption{Adaptation to commercial online generators---top: Doubao (text ``Support Skywriters'' in smoke trails), bottom: GPT Image (a dining table above a bear). Each cell shows the rendered image with its prompt beneath. TARA produces the complete smoke-trail text and places the table directly above the bear, while the baselines leave the text incomplete or split the two objects.}
\label{fig:crossgen}
\end{figure}

Beyond the open-source generators studied above, we check whether TARA's typed prompts transfer to commercial online systems. Figure~\ref{fig:crossgen} compares the original, VisualPrompter, and TARA prompts on two such systems---Doubao on a smoke-trail text case, and GPT Image on a ``dining table above a bear'' spatial case. In both, TARA's typed rewrite realizes the layout the prompt asks for---a complete smoke-trail script with a visible contrail, and a table placed directly above the bear---whereas the baselines leave the text incomplete or split the two objects onto separate levels.

\subsection{Failure Cases}
\label{app:failures}
\begin{figure}[htbp]
\centering
\includegraphics[width=\textwidth]{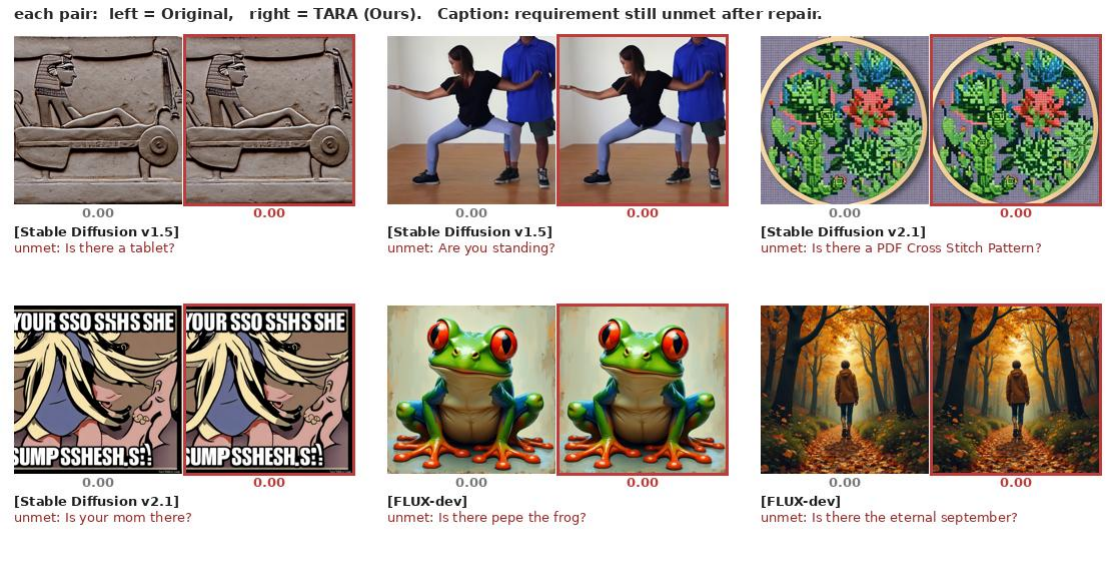}
\caption{Failure cases (DSG). Left: original; right (red outline): TARA. Numbers are VLM semantic scores; the caption under each pair is a requirement still unmet after repair.}
\label{fig:failures}
\end{figure}

Figure~\ref{fig:failures} shows cases where the optimized image still misses a requirement. Most residual failures stem from generator-side limits---multi-object counting, rare poses, and long rendered strings---where a better prompt is not enough to force compliance; the repair gate then keeps the higher-scoring of the two images under the atomic evaluator but cannot manufacture a capability the generator lacks.

\section{Supplementary Quantitative Results}

\subsection{Behavioral Analysis}
\label{app:behavior}

\begin{figure}[htbp]\centering
\includegraphics[width=0.6\textwidth]{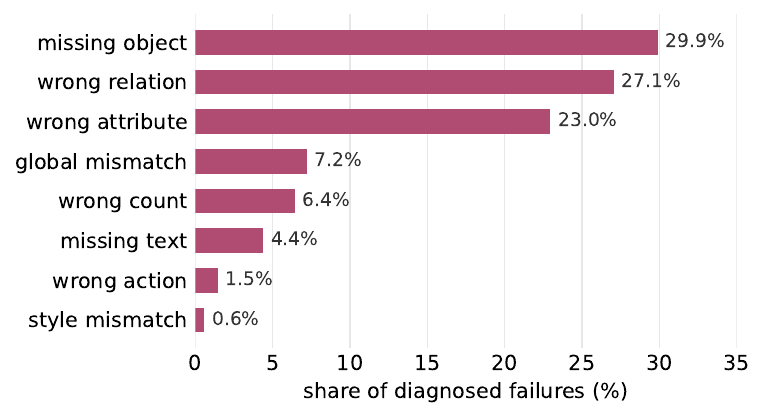}
\caption{Distribution of all eight diagnosed error types over failed atomic propositions, aggregated across the 24 benchmark--generator--seed cells.}
\label{fig:errtypes}
\end{figure}

Diagnosed failures are diverse. Figure~\ref{fig:errtypes} shows the complete eight-type distribution over all diagnosed failures. Missing objects ($29.9\%$), wrong relations ($27.1\%$), and wrong attributes ($23.0\%$) are most frequent, while action and style errors are rarer but remain explicitly routed by TARA. A single uniform expansion cannot match this heterogeneity; routing each failure to a type-specific repair is exactly what this calls for.

\begin{figure}[htbp]\centering
\includegraphics[width=0.6\textwidth]{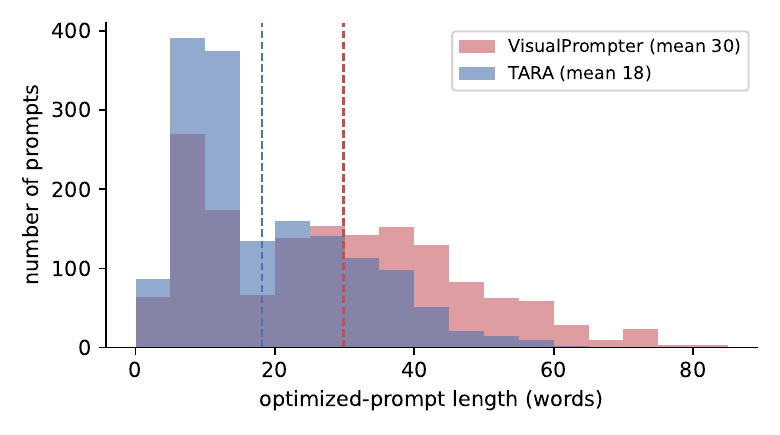}
\caption{Optimized-prompt length over all prompts (DSG and TIFA, four generators). TARA's compiled typed repairs are substantially shorter than VisualPrompter's uniform expansions.}
\label{fig:promptlen}
\end{figure}

TARA's prompts stay short. Figure~\ref{fig:promptlen} compares the length of the optimized prompts. Over all prompts TARA averages $18$ words versus $30$ for VisualPrompter, because typed repair compilation makes a short targeted edit instead of a verbose global expansion.

Table~\ref{tab:promptlen_all} extends the length comparison to all seven methods. Aesthetics-oriented optimizers inflate the prompt the most (NeuroPrompts and TIPO exceed $50$ words), VisualPrompter's uniform expansion averages $30$, whereas TARA's typed repair compilation stays at $18$---the shortest among the rewriters while remaining the most faithful.

\begin{table}[htbp]
\centering
\small
\caption{Optimized-prompt length (mean words) by method and generator, over all prompts. Aesthetics-oriented optimizers inflate the prompt the most; VisualPrompter's uniform expansion is long, whereas TARA's typed repair compilation stays the shortest among the rewriters while still being the most faithful (cf.\ main results).}
\label{tab:promptlen_all}
\begin{tabular}{l cccc c}
\toprule
Method & SD 1.5 & SD 2.1 & Flux-dev & Janus-Pro & All \\
\midrule
Raw prompt & 14.2 & 14.2 & 14.2 & 14.2 & 14.2 \\
NeuroPrompts & 57.5 & 57.5 & 57.5 & 57.5 & 57.5 \\
Promptist & 23.1 & 23.1 & 23.1 & 23.1 & 23.1 \\
BeautifulPrompt & 29.1 & 29.1 & 29.1 & 29.1 & 29.1 \\
TIPO & 47.0 & 47.0 & 47.0 & 73.0 & 53.5 \\
\rowcolor{cSecond} VisualPrompter & 32.2 & 31.3 & 28.5 & 29.1 & 30.3 \\
\rowcolor{cBest} TARA (Ours) & 18.3 & 18.0 & 18.2 & 18.5 & 18.2 \\
\bottomrule
\end{tabular}
\end{table}

The repair traces let us inspect TARA's behavior per failure type. Table~\ref{tab:mech_errortype} reports, for each diagnosed error type, its frequency, how often a typed rewrite improves the semantic score and is kept, and the mean gain when accepted: missing objects and actions are the most repairable, whereas rendered text is the hardest, echoing the per-type accuracy in the main text. Table~\ref{tab:mech_etxgen} breaks the accept rate down by generator---the same failure type is far more repairable on stronger generators, indicating that part of the ceiling is generator capacity rather than the optimizer.

\begin{table}[htbp]
\centering
\small
\caption{Repair behavior by error type (TARA, aggregated over all 24 benchmark--generator--seed cells). \emph{Frequency} is the number of diagnosed failures of each type; \emph{accept rate} is the fraction whose typed rewrite improved the semantic score and was kept; \emph{mean gain} is the average score increase of accepted repairs. Object and action failures are the most repairable, whereas rendered text is the hardest---consistent with the per-type accuracy in the main text. Each diagnosed failure occurrence inherits its prompt's gate decision, so prompts containing multiple error types contribute to multiple rows.}
\label{tab:mech_errortype}
\begin{tabular}{l r c c}
\toprule
Error type & Frequency & Accept rate (\%) & Mean gain \\
\midrule
Missing object & 4837 & 29.7 & $+0.133$ \\
Wrong relation & 4379 & 27.7 & $+0.084$ \\
Wrong attribute & 3716 & 24.5 & $+0.061$ \\
Wrong count & 1042 & 27.0 & $+0.094$ \\
Wrong action & 238 & 29.0 & $+0.114$ \\
Global mismatch & 1168 & 26.5 & $+0.075$ \\
Missing text & 711 & 22.5 & $+0.048$ \\
Style mismatch & 95 & 26.3 & $+0.080$ \\
\bottomrule
\end{tabular}
\end{table}

\begin{table}[htbp]
\centering
\small
\caption{Repair accept rate (\%) by error type and generator (TARA). Darker is higher. The same failure type is far more repairable on stronger generators---e.g.\ rendered text is nearly unrepairable on Stable Diffusion but much easier on Flux/Janus---showing that part of the ceiling is generator capacity, not the optimizer.}
\label{tab:mech_etxgen}
\begin{tabular}{l cccc}
\toprule
Error type & SD 1.5 & SD 2.1 & Flux-dev & Janus-Pro \\
\midrule
Missing object & \cellcolor{cPos!46}29 & \cellcolor{cPos!47}29 & \cellcolor{cPos!49}31 & \cellcolor{cPos!50}31 \\
Wrong relation & \cellcolor{cPos!43}27 & \cellcolor{cPos!43}27 & \cellcolor{cPos!48}30 & \cellcolor{cPos!44}28 \\
Wrong attribute & \cellcolor{cPos!39}24 & \cellcolor{cPos!39}24 & \cellcolor{cPos!37}23 & \cellcolor{cPos!42}26 \\
Wrong count & \cellcolor{cPos!42}26 & \cellcolor{cPos!43}27 & \cellcolor{cPos!44}28 & \cellcolor{cPos!44}27 \\
Wrong action & \cellcolor{cPos!48}30 & \cellcolor{cPos!48}30 & \cellcolor{cPos!48}30 & \cellcolor{cPos!42}26 \\
Global mismatch & \cellcolor{cPos!39}24 & \cellcolor{cPos!45}28 & \cellcolor{cPos!42}26 & \cellcolor{cPos!43}27 \\
Missing text & \cellcolor{cPos!33}21 & \cellcolor{cPos!32}20 & \cellcolor{cPos!43}27 & \cellcolor{cPos!39}25 \\
Style mismatch & \cellcolor{cPos!32}20 & \cellcolor{cPos!32}20 & \cellcolor{cPos!59}37 & \cellcolor{cPos!40}25 \\
\bottomrule
\end{tabular}
\end{table}

\subsection{Threshold Sensitivity}
\label{app:ablation_sensitivity}

TARA fixes a single global $\tau{=}0.72$ (the same value across every generator and benchmark) to admit the conservative re-seed candidate only when the initial image is already near-correct. A threshold sweep on DSG/SD~2.1 (Table~\ref{tab:tau_sensitivity}) shows that all tested values in $[0.60,0.84]$ produce large gains over the initial image ($+8$ to $+10$ points) at a similar image budget ($1.78$--$1.81$ generations). We therefore keep $\tau{=}0.72$ fixed throughout, without tuning it per generator or benchmark.

\begin{table}[H]
\centering
\small
\caption{Sensitivity to the re-seed threshold on DSG/SD~2.1 (single seed $s{=}1234$, $n{=}200$). All tested thresholds give large gains over the initial image at a comparable image budget. The default row reuses the main-experiment run, whereas the remaining rows come from a separately launched sensitivity sweep. Although the nominal seed is shared, image generation is not guaranteed to be bitwise deterministic across independent GPU sessions, leading to the small difference in Init.}
\label{tab:tau_sensitivity}
\begin{tabular}{lcccc}
\toprule
Threshold $\tau$ & Init & Final & $\Delta$ & Gens \\
\midrule
0.60 & 65.48 & 73.63 & $+8.16$ & 1.78 \\
0.66 & 65.48 & 74.49 & $+9.01$ & 1.78 \\
\rowcolor{cSecond} 0.72 (default) & 65.06 & 73.69 & $+8.63$ & 1.81 \\
0.78 & 65.48 & 74.75 & $+9.28$ & 1.78 \\
0.84 & 65.48 & 75.41 & $+9.93$ & 1.78 \\
\bottomrule
\end{tabular}
\end{table}


\subsection{Robustness and Diagnostic Breakdowns}

To quantify uncertainty in the comparison with VisualPrompter, we use a prompt-clustered paired bootstrap. Within each benchmark, one cluster contains a prompt's 12 paired TARA-minus-VisualPrompter differences across four generators and three seeds; we average within each cluster and resample the 200 prompt clusters with replacement for 100{,}000 replicates (random seed 20260713). The overall interval averages benchmark-stratified bootstrap draws. The 2.5th and 97.5th percentiles give 95\% CIs of $[4.72,6.52]$ for DSG, $[1.63,3.63]$ for TIFA, and $[3.45,4.80]$ overall, all excluding zero.

Table~\ref{tab:appendix_seed} reports semantic accuracy by seed group, each averaged over the four generators; the scores are stable across the three groups, confirming the main results are not seed-specific.

\begin{table}[H]
\centering
\scriptsize
\caption{Random-seed robustness. Each seed column averages semantic accuracy (\%) over four generators; the final column reports mean $\pm$ standard deviation over the three seed groups.}
\label{tab:appendix_seed}
\resizebox{\textwidth}{!}{%
\begin{tabular}{l cccc cccc}
\toprule
\multirow{2}{*}{Method} & \multicolumn{4}{c}{DSG} & \multicolumn{4}{c}{TIFA} \\
\cmidrule(lr){2-5}\cmidrule(lr){6-9}
& s1234 & s2345 & s3456 & Mean $\pm$ Std & s1234 & s2345 & s3456 & Mean $\pm$ Std \\
\midrule
Raw prompt & 68.4 & 69.0 & 67.3 & 68.3 $\pm$ 0.9 & 76.4 & 75.8 & 76.5 & 76.2 $\pm$ 0.4 \\
NeuroPrompts & 62.4 & 62.1 & 62.7 & 62.4 $\pm$ 0.3 & 68.8 & 70.1 & 68.0 & 69.0 $\pm$ 1.0 \\
Promptist & 65.0 & 64.9 & 64.0 & 64.6 $\pm$ 0.6 & 74.0 & 74.6 & 73.7 & 74.1 $\pm$ 0.4 \\
BeautifulPrompt & 41.8 & 42.8 & 42.0 & 42.2 $\pm$ 0.6 & 50.3 & 49.6 & 48.8 & 49.6 $\pm$ 0.7 \\
TIPO & 58.4 & 59.9 & 58.3 & 58.9 $\pm$ 0.9 & 60.8 & 60.4 & 60.8 & 60.7 $\pm$ 0.2 \\
VisualPrompter & 71.5 & 70.3 & 70.5 & 70.8 $\pm$ 0.7 & 82.5 & 82.4 & 83.4 & 82.8 $\pm$ 0.6 \\
TARA (Ours) & 76.2 & 76.6 & 76.3 & 76.4 $\pm$ 0.2 & 85.1 & 85.5 & 85.4 & 85.4 $\pm$ 0.2 \\
\bottomrule
\end{tabular}}
\end{table}


\subsection{VLM-as-Judge Preference Details}\label{app:pref}

Table~\ref{tab:appendix_pref} details the VLM-as-Judge preference between TARA and the raw prompt on changed-prompt cases, for both semantic consistency and aesthetics.

\begin{table}[H]
\centering
\scriptsize
\caption{Detailed VLM-as-Judge preference on changed-prompt cases. Each entry is TARA / Tie / Raw preference percentage.}
\label{tab:appendix_pref}
\begin{tabular}{l c c c}
\toprule
Model & $n$ & Semantic Consistency & Aesthetics \\
\midrule
SD 1.5 & 17 & 70.6 / 17.6 / 11.8 & 52.9 / 0.0 / 47.1 \\
Flux-dev & 12 & 66.7 / 0.0 / 33.3 & 50.0 / 0.0 / 50.0 \\
\midrule
Average & 29 & 69.0 / 10.3 / 20.7 & 51.7 / 0.0 / 48.3 \\
\bottomrule
\end{tabular}
\end{table}

\end{document}